\documentclass{article}

\usepackage{url}

\usepackage[english]{babel}

\usepackage[utf8]{inputenc}
\usepackage{xspace}
\usepackage{amsmath,amsthm,amssymb,mathtools}
\usepackage{lmodern,algorithm}

\usepackage[algo2e,ruled,vlined,linesnumbered]{algorithm2e}

\usepackage{xcolor}
\usepackage{tikz}
\usepackage{tikz} 
\usetikzlibrary{shapes,
  decorations,
  arrows,
  patterns,
  positioning,
  plotmarks,
  calc,
  intersections,
  decorations.pathmorphing,
  decorations.markings}

\usepackage{graphicx}

\allowdisplaybreaks[4]
\newtheorem{theorem}{Theorem}
\newtheorem{lemma}[theorem]{Lemma}

\newtheorem{definition}[theorem]{Definition}
\newtheorem{proposition}[theorem]{Proposition}

\usepackage{dsfont}

\newcommand{\OPT}{\mathsf{OPT}}
\newcommand{\gsemo}{Global SEMO\xspace}

\newcommand{\gsemoalt}{Global SEMO$_{alt}$\xspace}
\def\cI{\ensuremath{\mathcal{I}}}

\newcommand{\Inn}[1]{\mathsf{Inn}\!\left({#1}\right)}
\newcommand{\Out}[1]{\mathsf{Out}\!\left({#1}\right)}
\newcommand{\inv}[2]{\sigma^{\rm I}_{#1#2}}  %% inversion operator
\newcommand{\jmp}[2]{\sigma^{\rm J}_{#1#2}}  %% jump operator
\def\fn{\ensuremath{A}}

\newcommand{\Tlopt}{$T_{lopt}$\xspace}
\newcommand{\Topt}{$T_{opt}$\xspace}
\newcommand{\Glopt}{$G_{loc}$\xspace}
\def\W1{\mathsf{W}[1]}

\newcommand{\GEA}{Generic (1+1)~EA\xspace}
\newcommand{\TEA}{Tree-Based (1+1)~EA\xspace}

\DeclareMathOperator{\poly}{poly}

\usepackage[textsize=footnotesize]{todonotes}

\usepackage{url}

\usepackage[english]{babel}

\usepackage[utf8]{inputenc}
\usepackage{xspace}
\usepackage{amsmath,amssymb,mathtools}
\usepackage{lmodern}

\usepackage[algo2e,ruled,vlined,linesnumbered]{algorithm2e}

\usepackage{xcolor}
\usepackage{tikz}
\usepackage{graphicx}

\allowdisplaybreaks[4]
\newcommand{\oea}{$(1 + 1)$~EA\xspace}

\newcommand{\mpoea}{$(\mu+1)$~EA\xspace}

\newcommand{\om}{\textsc{OneMax}\xspace}

\newcommand{\R}{\ensuremath{\mathbb{R}}}

\newcommand{\N}{\ensuremath{\mathbb{N}}}

\usepackage{authblk}

\newcommand{\mpleak}{$(\mu+\lambda)$~EA$^k$\xspace}
\newcommand{\mplea}{$(\mu+\lambda)$~EA\xspace}

\title{Parameterized Complexity Analysis of\\ Randomized Search Heuristics}
\author[1]{Frank Neumann}
\author[2]{Andrew M. Sutton}
\affil[1]{Optimisation and Logistics Group, School of Computer Science, The University of Adelaide, Australia}  
\affil[2]{Department of Computer Science, University of Minnesota Duluth, USA}
\date{}

\usepackage{fullpage}
\begin{document}
\maketitle

\begin{abstract}
  This chapter compiles a number of results that apply the theory of parameterized algorithmics to the running-time analysis of randomized search heuristics such as evolutionary algorithms. The parameterized approach articulates the running time of algorithms solving combinatorial problems in finer detail than traditional approaches from classical complexity theory. We outline the main results and proof techniques for a collection of randomized search heuristics tasked to solve $\mathsf{NP}$-hard combinatorial optimization problems such as finding a minimum vertex cover in a graph, finding a maximum leaf spanning tree in a graph, and the traveling salesperson problem.
\end{abstract}

\numberwithin{equation}{section}

\section{Introduction}
\label{sec:paramcomplex:intro}

Randomized search heuristics (RSHs) are a class of general-purpose algorithms that are often deployed to tackle hard combinatorial optimization problems that arise in practice. 
Instances of practical, real-world problems are usually structured or restricted in some way, and it is typically assumed that RSH techniques are successful when the underlying strategy is able to exploit the structural properties of the resulting search space. 

The mathematical analysis of the running time of randomized search heuristics on discrete optimization problems has advanced in the last decade. For a wide array of these techniques, rigorous and precise asymptotic bounds on the performance as a function of problem size are now available. 
However, many of these kinds of results are restricted only to toy problems. While such analyses are useful for gaining an understanding of the general working principles underlying RSH techniques, it is often not clear how they might be interpreted in the context of classically hard problems in computer science. 

Unless $\mathsf{P} = \mathsf{NP}$, the worst-case runtime of an \textsf{NP}-hard problem cannot be bounded from above by a polynomial in the input size. This is a rather restrictive view, and it often tells us nothing about the typical behavior of algorithms on problems that are likely to be encountered in practice. For example, many experimental studies confirm that randomized search heuristics such as evolutionary algorithms (EAs), ant colony optimization, simulated annealing, and simple hill-climbing perform well on practical instances of \textsf{NP}-hard problems. An important research question for RSH techniques applied to combinatorial optimization is: which features of a given instance determine its hardness, and how do such parameters influence the runtime?

The field of \emph{parameterized complexity} offers a refinement of classical time complexity by analyzing the running time of an algorithm not just as a function of problem size, but also as a function of further parameters of the input, for example, solution size, structural restrictions, or quality of approximation~\cite{RodFell1999,Flum2006parameterized}. The idea is to capture the essence of what makes a problem instance hard, and try to isolate this hardness to some structural feature of the instance or its solution. The inevitable combinatorial explosion in the runtime is confined to a function of this parameter, with only polynomial dependence on the input size. Even large instances may exhibit a very restricted structure and can be easier to solve, independent of size. Parameterized complexity is therefore an obvious candidate for systematically studying what features of a particular problem are hard for RSH techniques. It can also offer advice on what types of problem might be soluble or insoluble by such approaches, and guide algorithm design. It should be noted that parameterized analysis can also be applied to study the efficiency of modules of an evolutionary algorithm. A good example is the hypervolume indicator, which has been widely applied in the area of evolutionary multiobjective optimization. Computing the optimal hypervolume is hard when the dimension grows, and the computation of the hypervolume has been investigated in \cite{DBLP:conf/gecco/BringmannF13} from a parameterized and average-case perspective.

Many hard problems have ``easy parts'' that can be efficiently solved in order to effectively shrink a problem to its computationally hard core structure. This can be done by efficiently reducing the problem instance to a smaller instance (kernelization), or constraining the search tree to a manageable size that is still guaranteed to contain a solution (bounded search tree method). A slower exact algorithm (even brute-force search) can then be run on the resulting smaller instance or search space.
With little to no hope of a polynomial-time solution, one instead seeks algorithms that can solve a problem in time that grows polynomially with the problem size, although perhaps superpolynomially with respect to some instance parameter. In other words, if the parameter is fixed to be small, the problem class is tractable, even as its instances grow large. Such a problem class (and corresponding algorithm) is called \emph{fixed-parameter tractable} (FPT). A slightly less desirable situation is an algorithm that runs in so-called \emph{slicewise polynomial time} (XP). Here the runtime is a polynomial in the problem size, but a polynomial whose degree depends on the parameter.

This kind of demarcation into hard and easy components can also be useful for the analysis of RSH techniques.
At the extreme end of the spectrum are functions such as \textsc{Needle}, whose black-box complexity establishes that no RSH could even beat simple random sampling in expectation. At the other extreme are problems from the \om class that are solved efficiently by even very simple approaches.
Likely, practical optimization problems lie somewhere between these two extremes, containing some mixture of components that can be efficiently exploited by randomized search heuristics and components that essentially require random sampling. If the hard core component that demands random sampling is guaranteed to be small by the nature of the problem class, then RSH techniques can be a reasonable approach. 
The theory of parameterized complexity is therefore useful for isolating the structural features that can be efficiently exploited by RSH techniques from the hard ``core'' of a problem, on which an approach must resort to some kind of stochastic brute-force search behavior such as random walks, lucky jumps, or explicit restarts. 

It should therefore not come as a surprise that analyzing randomized search heuristics from the perspective of parameterized complexity can lead to useful theoretical insights into algorithm design. For example, it has been shown that the specific choice of search operator can directly influence the fixed-parameter tractability of an algorithm on certain problems, for example, tree-preserving mutation on the maximum-leaf spanning tree problem~\cite{DBLP:conf/ppsn/KratschLNO10} or standard uniform crossover on the closest-string problem~\cite{DBLP:conf/gecco/Sutton18}.

The aim of this chapter is to discuss a number of results in the field of parameterized complexity applied to RSH techniques. We begin in Section~\ref{sec:paramcomplex:param} by introducing some background and technical details. In Section~\ref{sec:paramcomplex:ml}, we consider the maximum-leaf spanning tree problem and show that the use of a mutation operator commonly used for spanning trees reduces the XP~runtime to FPT~runtime when compared with standard bit mutations.
In Section~\ref{sec:paramcomplex:vc}, we discuss multiobjective evolutionary algorithms that quickly focus their search on a kernel of minimum vertex cover instances, and subsequently perform random sampling on that kernel, resulting in FPT~runtime.
Decomposing the runtime analysis of an algorithm into a set of instance parameters is useful in its own right to better understand the components of a problem that influence the behavior of search heuristics. In Section~\ref{sec:paramcomplex:sub}, we present results on the maximization of submodular functions under different constraints. These results derive the expected time that simple evolutionary algorithms need to produce approximations as a function of both the problem size and additional parameters of the input. In Section~\ref{sec:paramcomplex:tsp}, we describe the analysis of a standard evolutionary algorithm (EA) applied to the Euclidean  traveling salesperson problem (TSP), which bounds the running time in the context of a well-known TSP parameterization (the number of points interior to the convex hull). In this case, it is possible to prove that the performance of the algorithm is bounded by the number of interior points, although this is not enough to obtain the desired fixed-parameter tractable runtime. On the other hand, if the EA is allowed to use some problem-specific information (namely, the cyclic order of points as they appear on the convex hull), it can explicitly focus its search on a small subset of states. This dramatic search space reduction yields fixed-parameter tractable runtimes for algorithms on parameterized TSP instances. We summarize the chapter in Section~\ref{sec:paramcomplex:conclusion} and briefly discuss some open research problems.

\section{Parameterized Complexity Analysis}
\label{sec:paramcomplex:param}
Extending traditional runtime analysis by parameterization requires conducting a rigorous runtime analysis of an algorithm on a \emph{parameterization} of a problem class. A parameterization of a problem class is a mapping of problem instances into the set of natural numbers. The running time of the algorithm is then expressed in terms of both the problem size and this extra parameter. 

Let $L$ be a language over a finite alphabet $\Sigma$. A \emph{parameterization} of $L$ is a mapping $\kappa : \Sigma^* \to \N$. The corresponding \emph{parameterized problem} is the pair $(L,\kappa)$. For a string $x \in \Sigma^*$, let $k = \kappa(x)$ and $n = |x|$. An algorithm deciding $x \in L$ in time bounded by $n^{g(k)}$ is called a \emph{slicewise polynomial-time} algorithm (or XP~algorithm). Here, $g : \N \to \N$ is an arbitrary but computable function. An algorithm deciding $x \in L$ in time bounded by $g(k) \cdot n^{O(1)}$ is called a \emph{fixed-parameter tractable} (or FPT) algorithm for the parameterization $\kappa$. Both kinds of algorithms
run in polynomial time for fixed $k$, but an XP~algorithm allows the degree of the polynomial to depend on the parameter, while the degree of the polynomial for the running time is independent of both $n$ and $k$ for an FPT~algorithm.

Randomized search heuristics are typically stochastic processes that are allowed to run for a certain number of iterations, after which the best-so-far result is collected and returned.  In each iteration, the process keeps a set of one or more candidate solutions, and evaluates their quality via a \emph{fitness} or \emph{objective} function. The candidate solutions for the next iteration are then computed using a number of transformation operations.

To analyze this class of algorithm, we consider a random variable $T$ that measures the number of basic iterations (usually measured in calls to the objective function) until a solution is first discovered. 
Here, a solution may be, depending on the context, an element that maximizes or minimizes the objective function. This allows us to treat optimization problems in the same manner as one would treat decision problems. 
Specifically, given a class of instances of an optimization problem, for each $N$ one can construct a decision problem $L \subseteq \Sigma^*$ as the set of all instances on which the maximum (or, minimum) objective function value is at least (or, at most) a particular value. 

The quantity $E[T]$ is the \textit{expected optimization time}, and is the most commonly used performance measure in the rigorous runtime analysis of randomized search heuristics.
We say an algorithm is a \emph{Monte~Carlo FPT~algorithm} for a parameterized problem $(L,\kappa)$ if it accepts $x \in L$ with probability at least $1/2$ in time $g(\kappa(x)) \cdot |x|^{O(1)}$ and accepts $x \not \in L$ with probability zero. Thus, any randomized search heuristic with a bound $E[T] \leq g(\kappa(x)) \cdot |x|^{O(1)}$ on $L$ can be trivially transformed into a Monte~Carlo FPT~algorithm by stopping its execution after $2g(\kappa(x))\cdot |x|^{O(1)}$ iterations. 

Note that the parameter is allowed to depend on the input in more or less an arbitrary way. The selection of a \emph{meaningful} parameterization depends strongly on what a ``typical'' problem instance looks like. In most cases, one hopes to choose a parameter that is assumed to be small over the set of problems one wishes to solve. Ideally, the parameter should somehow capture the source of exponential complexity for the problem~\cite{Flum2006parameterized}. 

The goal of applying parameterized complexity analysis to the field of randomized search heuristics is thus to somehow understand how much information from the fitness function can be exploited in more detail. At the worst extreme, there is no exploitable information in the fitness of solutions at all (i.e., the fitness of a solution tells us nothing about its relationship to a global optimum), and we are in a blind \textsc{Needle}-like case. Any RSH technique that employs such a fitness function must then rely entirely on getting lucky enough to stumble on an optimal solution. However, as previously mentioned, for most realistic problems we conjecture that there exists some structure in the fitness function that can be implicitly used by the RSH technique. Parameterized analysis can be seen as a technique that allows us to inspect the fitness function to assist in bounding how much ``luck'' is required to solve the problem.

\section{Maximum-Leaf Spanning Trees}
\label{sec:paramcomplex:ml}
The classical minimum spanning tree problem, which can be solved in polynomial time by well-known deterministic algorithms such as those of Kruskal and Prim, has gained significant attention in the evolutionary computation literature~\cite{NeumannWegenerTCS07,algorithmica/DoerrJW12}. This includes the investigations of Witt~\cite{DBLP:conf/gecco/Witt14}, who considered an additional structural parameter of the given graph. He gave an upper bound on the runtime of simple evolutionary algorithms for the minimum spanning tree problem that depends on the circumference of the given graph. We will not present the details here, as the focus of this chapter is on \textsf{NP}-hard problems. We instead refer the interested reader to the original articles.

We start our investigations by considering an \textsf{NP}-hard variant of a spanning tree problem where the choice of mutation operator
affects the parameterized runtime. Specifically, the commonly used standard bit mutation operation results in XP~runtime, whereas a mutation operator that creates feasible solutions produces FPT~runtime.

The problem we consider is the maximum-leaf spanning tree problem, and we summarize the results given in \cite{DBLP:conf/ppsn/KratschLNO10}. 
Given an undirected, connected graph $G=(V,E)$, the goal is to find a spanning tree $T^*$ of $G$ such that the number of leaves is maximum.

The authors of~\cite{DBLP:conf/ppsn/KratschLNO10} considered two simple evolutionary algorithms that differ in the choice of the mutation operator. The first algorithm uses a general mutation operator carrying out standard bit mutations, and the second is specific to spanning tree problems.
Both algorithms start with an arbitrary spanning tree $T$ of $G$. We
denote by $m$ the number of edges in $G$, and by $\ell(T)$ the number of leaves of the spanning tree $T$. A new solution is accepted only if it is a spanning tree whose number of leaves is at least as high as the number of leaves in the current solution.
The algorithm called the \GEA is given in Algorithm~\ref{alg:paramcomplex:flip}.

\begin{algorithm2e}
  \caption{\label{alg:paramcomplex:flip} \GEA}
  \SetKwFor{For}{repeat}{}{}
  \SetKw{And}{and}
  Choose a spanning tree of $T$ uniformly at random\;
  \For{forever}{%
    Produce $T'$ by swapping each edge of $T$ independently with probability $1/m$\;
    \lIf{$T'$ is a tree \And $\ell(T') \geq \ell(T)$}{$T\gets T'$}
}
\end{algorithm2e}

Swapping an edge in the mutation step of the \GEA means that if an edge is present in $T$ then it is not contained in $T'$ with probability $1/m$. On the other hand, if an edge is not present in $T$ then it is contained in $T'$ with probability $1/m$. An edge does not change from $T$ to $T'$ with probability $1-1/m$ in each mutation step, independently of the other edges.

The mutation operator of Algorithm~\ref{alg:paramcomplex:flip} does not necessarily create an offspring that is a tree. If the offspring is not a tree, then this individual is discarded, as it represents an infeasible solution.

The second algorithm we consider is called the \TEA and is illustrated in Algorithm~\ref{alg:paramcomplex:edgeexchange}. This approach uses
a problem-specific mutation operator that ensures valid solutions, i.e., spanning trees. It is well known that, given a spanning tree $T$, a new spanning tree $T'$ can be created by introducing an edge $e \in E \setminus T$ and removing an edge from the resulting cycle. Mutation operators based on this idea are commonly used when applying evolutionary algorithms to \textsf{NP}-hard spanning tree problems.

\begin{algorithm2e}
  \caption{\label{alg:paramcomplex:edgeexchange}\TEA}  
  \SetKwFor{For}{repeat}{}{}
  Choose an arbitrary spanning tree $T$ of $G$\;
  \For{forever}{%
    Choose $S$ according to a Poisson distribution with parameter $\lambda=1$ and perform sequentially $S$ random edge-exchange operations to obtain a spanning tree $T'$. A random exchange operation applied to a spanning tree $\tilde{T}$ chooses an edge $e \in E \setminus \tilde{T}$ uniformly at random. The edge $e$ is inserted and one randomly chosen edge of the cycle in $\tilde{T} \cup \{e\}$ is deleted\;
    \lIf{$\ell(T') \geq \ell(T)$}{$T\gets T'$}
  }
\end{algorithm2e}

Our goal is to point out the differences between the two algorithms.
To do this, we compare the expected optimization time $E[T]$ of the two algorithms. This shows that the problem-specific mutation operator of Algorithm~\ref{alg:paramcomplex:edgeexchange} makes the difference between a fixed-parameter evolutionary algorithm and an evolutionary algorithm that cannot compute an optimal solution in expected FPT~time.

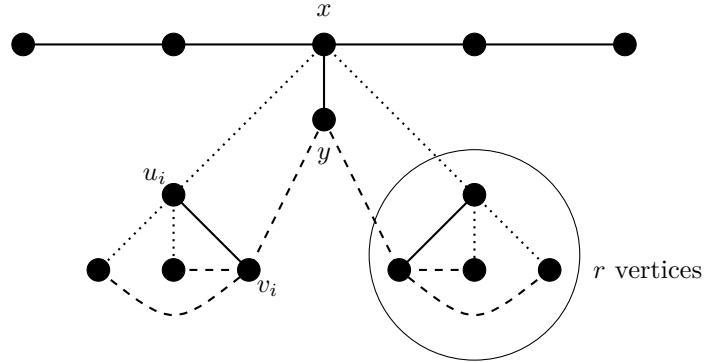
\begin{figure}[tb]
    \centering
\begin{tikzpicture}

\draw[thick] (0,0) -- (2,0) -- (4,0) -- (6,0) -- (8,0);
\draw[thick] (4,0) -- (4,-1);
\draw[thick] (2,-2) -- (3,-3);
\draw[thick] (6,-2) -- (5,-3);

\draw[dotted,thick] (4,0) -- (2,-2) -- (1,-3);
\draw[dotted,thick] (2,-2) -- (2,-3);
\draw[dotted,thick] (4,0) -- (6,-2) -- (6,-3);
\draw[dotted,thick] (6,-2) -- (7,-3);

\draw[dashed,thick] (2,-3) -- (3,-3) -- (4,-1);
\draw[dashed,thick] (6,-3) -- (5,-3) -- (4,-1);
\draw[dashed,thick] (1,-3) .. controls (2,-3.8) .. (3,-3);
\draw[dashed,thick] (7,-3) .. controls (6,-3.8) .. (5,-3);

\draw (6,-2.8) circle (1.4cm);
\draw (8.3,-3) node {$r$ vertices};

\draw[fill=black] (0,0) circle (0.15cm);
\draw[fill=black] (2,0) circle (0.15cm);
\draw[fill=black] (4,0) circle (0.15cm) node[above=0.25cm] {$x$};
\draw[fill=black] (6,0) circle (0.15cm);
\draw[fill=black] (8,0) circle (0.15cm);
\draw[fill=black] (4,-1) circle (0.15cm) node[below=0.25cm] {$y$};
\draw[fill=black] (2,-2) circle (0.15cm);
\draw[fill=black] (6,-2) circle (0.15cm);
\draw[fill=black] (1,-3) circle (0.15cm);
\draw[fill=black] (2,-3) circle (0.15cm);
\draw[fill=black] (3,-3) circle (0.15cm);
\draw[fill=black] (5,-3) circle (0.15cm);
\draw[fill=black] (6,-3) circle (0.15cm);
\draw[fill=black] (7,-3) circle (0.15cm);

\draw (1.75,-1.75) node {$u_i$};
\draw (3.25,-3.25) node {$v_i$};
\end{tikzpicture}
\caption{Local optimum, shown with dashed edges, and global optimum, shown with dotted edges; shared edges are drawn solid.}
    \label{fig:paramcomplex:lopt}
\end{figure}

For the \GEA, the authors of~\cite{DBLP:conf/ppsn/KratschLNO10} gave a lower bound which showed that the algorithm cannot solve the problem in FPT time. They considered the graph given in Fig.~\ref{fig:paramcomplex:lopt}.
The instance contains a local optimum, which has a distance to the global optimum in terms of the number of edges that have to be exchanged. The number of these edge exchanges depends on the number of nodes, $r$, the magnitude of which can be chosen to make it hard or easy to escape from the local optimum.

Formally, our graph, called \Glopt (see Fig.~\ref{fig:paramcomplex:lopt}) contains two components consisting of $r$ vertices each. In component $i$, $1 \leq i \leq 2$, two vertices $u_i$ and $v_i$  are connected to all the other vertices in that component. 
The vertex $u_i$ is connected to vertex $x$, which lies outside the component. Similarly, vertex $v_i$ is connected to vertex $y$. In addition, $x$ and $y$ share an edge. The graph is completed by attaching a path of $n-2r-2$ vertices to the vertex $x$.
A tree has to contain all the edges of the path attached to $x$.
In addition, at least one of the edges $\{u_i,x\}$ and $\{v_i, y\}$ has to be chosen for each $i$.
 For a given component, the maximum number of possible leaves is at most $r-1$. This can be obtained by attaching all nodes of the component either to $u_i$ or to $v_i$. 

The graph contains a local optimum \Tlopt which consists of all edges attached to the vertices $v_i$, $1 \leq i \leq 2$, the edge $\{x,y\}$, and all path edges. The global optimum \Topt consists of all edges attached to the vertices $u_i$, $1 \leq i \leq 2$, the edge $\{x,y\}$, and all path edges. 
Compared with \Tlopt, \Topt has an extra leaf, namely the vertex $y$.
However, \Tlopt and \Topt differ by $4(r-1)$, edges which make it hard for the algorithms under consideration to obtain \Topt if \Tlopt has been produced before.

\Tlopt can only by improved by swapping at least $2(r-2)$ edges, as all nonsolid edges adjacent to at least one node $v_i$ need to be swapped to reach an improvement.
As each bit corresponding to an edge of the graph is flipped with probability $1/m$ in the \GEA, the following lower bound on the expected optimization time of the \GEA is obtained.

\begin{theorem}
\label{thm:paramcomplex:lbGEA}
The expected optimization time of the \GEA on \Glopt is  lower bounded by  $\left(m/c\right)^{2(r-2)}$ where $c$ is an appropriate constant.
\end{theorem}

Using the same arguments, a lower bound of  $\left((r-2)/c\right)^{r-2}$ where $c$ is an appropriate constant, has been given for the \TEA. Again the bound considers the time to improve the locally optimal solution, which requires $r-2$ edge exchanges. The mutation operator of the \TEA has the benefit that a spanning tree is always created by introducing an edge and removing an edge from the resulting cycle, which results in a lower bound that is smaller than the one obtained for the \GEA. In terms of upper bounds, the \TEA runs in FPT time when the value of an optimal solution $k$ is the parameter.

The proof of the main result builds on the following lemma, which upper bounds the number of edges and the number of nodes of degree at least three as a function of $k$.

\begin{lemma}\label{lem:paramcomplex:edgebound}
Any connected graph~$G$ on~$n$ nodes and with a maximum number of~$k$ leaves in any spanning tree has at most~$n+5k^2-7k$ edges and at most~$10k-14$ nodes of degree at least three.
\end{lemma}

Each spanning tree has $n-1$ edges, which implies that the number of edge exchanges to obtain a maximum-leaf spanning tree from any spanning tree is $n+5k^2-7k - (n-1)\leq 5 k^2$. Furthermore, a nonoptimal spanning tree can be improved by removing an edge of degree two from the cycle.
The number of nodes of degree at least $3$ is at most $10k-14$,  which gives a lower bound of $1/20k$ on the probability of removing an edge of degree two from the cycle.

The upper bound for the \TEA is given in the following theorem, and the proof uses the arguments stated above.

\begin{theorem} \label{thm:paramcomplex:FPT-theorem}
  If the maximum number of leaf nodes in any spanning tree of $G$ is
  $k$, then the \TEA finds an optimal solution
  in expected time $O(2^{15k^2\log k})$.
\end{theorem}

\section{Minimum Vertex Cover}
\label{sec:paramcomplex:vc}

The minimum vertex cover problem is an important classical \textsf{NP}-hard combinatorial optimization problem. Given an undirected connected graph $G=(V,E)$, the task is to find a minimum set of vertices $V' \subseteq V$ such that each edge $e\in E$ is covered by one of the chosen nodes, i.e., $e\cap V' \not = \emptyset$ holds for each $e \in E$. A set of vertices $V'$ covering each edge $e \in E$ is called a vertex cover.

Using a binary variable $x_i$ for each vertex $v_i \in V$, the minimum vertex cover problem can be formulated as the following integer linear program (ILP):
\begin{equation*}
\begin{aligned}
  &\text{minimize}  & &\displaystyle\sum\limits_{i=1}^{n} x_{i}&\\
  &\text{subject to}& &x_i + x_j \geq 1,  & \forall \{i,j\} \in E,\\
  &                 & &x_i \in \{0,1\},   & 1 \leq i \leq n.\\
\end{aligned}
\end{equation*}
The linear program (LP) relaxation is obtained by relaxing the requirement $x_i \in \{0,1\}$ to $x_i \in [0,1], 1 \leq i \leq n$

The vertex cover problem is the most prominent problem in the area of parameterized complexity. As stated before, this area usually deals with decision problems. In the case of the vertex cover problem, one asks whether a given graph $G$ has a vertex cover of at most $k$ nodes.

Earlier studies~\cite{ECJ2,DBLP:journals/tec/OlivetoHY09} on the performance of the \oea have shown that this algorithm may get stuck in the smaller component of a complete bipartite graph when the two partitions have different sizes. Escaping this local optimum requires the algorithm to flip all bits belonging to the global optimum at once, and therefore has a waiting time of $\Omega(n^{\OPT})$, where $\OPT$ is the value of an optimal solution. Furthermore, if the two partitions $V_1$ and $V_2$ of the bipartite graph are extremely unbalanced, say $|V_1| = n^{\epsilon}$ and $|V_2| = n^{1-\epsilon}$, where $\epsilon>0$ is an arbitrary small constant, then the approximation ratio achieved by getting stuck in a local optimum is only $n^{1-\epsilon}/ n^{\epsilon} = n^{1- 2 \epsilon}$ and can therefore be made very close to the trivial approximation achieved by selecting all vertices of the given graph.

\subsection{\gsemo}
We consider the search space $\{0,1\}^n$, where each bit $x_i$ of a search point $x$ corresponds to a vertex~$v_i$ of the 
given graph $G$. The vertex~$v_i$ is chosen in the solution~$x$ iff $x_i=1$. 
The task is to find a solution~$x$ with a minimum number of vertices that covers all edges. This motivates us to introduce 
a fitness function based on the number of edges left uncovered by $x$.

We denote by $E(x)$ the set of edges covered by the cover $x$, i.e., $E(x) := \{e \mid e \cap V_x \not = \emptyset\}$, where $V_x := \{v_i \mid x_i=1, 1 \leq i \leq n\}$ is the subset of vertices chosen by $x$.

Kratsch and Neumann~\cite{DBLP:journals/algorithmica/KratschN13} considered two fitness functions for minimum vertex cover. 
The first fitness function was
\[
f_1(x) = (|x|_1, u(x)),
\]
where $|x|_1 = |\{i : x_i = 1\}|$ corresponds to the number of chosen vertices and $u(x) :=| E\setminus E(x)|$ is the number of edges left uncovered by $x$. Note that $u(x)$ is useful for directing the search process towards a feasible solution, i.e., a 
solution $x$ for which $u(x)=0$ holds. This function had already been considered in~\cite{ECJ2} in the context of approximations.

In addition, the authors of~\cite{DBLP:journals/algorithmica/KratschN13} examined a second fitness function that uses additional information obtained from a linear program. Let $G(x) = (V, E \setminus E(x))$ be the graph obtained from $G$ by removing all edges covered by nodes in $x$. We also consider the fitness function
\[
f_2(x) = (|x|_1, LP(x)),
\]
where~$LP(x)$ denotes the optimum value of the relaxed vertex cover ILP for $G(x)$, i.e., the cost of an optimal fractional vertex cover of~$G(x)$.

\begin{algorithm2e}
  \caption{\label{alg:paramcomplex:gsemo}\gsemo}  
  \SetKwFor{For}{repeat}{}{}
  Choose an initial solution $x\in\{0,1\}^n$ uniformly at random\;
  Determine $f(x)$ and initialize $P \gets \{x\}$\;
  \For{forever}{%
    Choose $x\in P$ randomly\;
    Create $x'$ by flipping each bit of $x$ independently with probability $1/n$\;
    Determine $f(x')$\;
    \If{$\exists x'' \in P,\, f(x'')\leq f(x')$ and $f(x'')\not= f(x')$}{$P$ is unchanged}
    \Else{exclude all $x''$ where $f(x')\leq f(x'')$ from $P$
      and add $x'$ to~$P$}
  }
\end{algorithm2e}

The multiobjective approach uses the \gsemo algorithm (see Algorithm~\ref{alg:paramcomplex:gsemo}). 
The algorithm starts with a bit string chosen uniformly at random. In each iteration, one individual $x$ of the current population $P$ is selected uniformly at random and undergoes standard bit mutation to produce an offspring $x'$. The offspring $x'$ is added to the population iff it is not strictly dominated by any other individual in $P$. In this case, all individuals in $P$ that are (weakly) dominated by $x'$ are removed from $P$.
We will examine \gsemo for the minimum vertex cover problem in this section and for maximization in several different types of problem involving submodular functions in the next section.

When minimizing the number of uncovered edges and the number of chosen vertices at the same time, \gsemo achieves an approximation to within a factor of $O(\log n)$ for the minimum vertex cover problem. These results may be generalized to the wider class of set cover problems. Kratsch and Neumann~\cite{DBLP:journals/algorithmica/KratschN13} have used a modification of \gsemo (called \gsemoalt) and shown that their approach computes an optimal solution in FPT time. 

\begin{algorithm2e}
  \caption{\label{alg:paramcomplex:ch-cover-altmutation}Alternative mutation operator in \gsemoalt}  
  \SetKw{Or}{or}
  Let $U(x)\subseteq E$ denote the set of 
  edges that are not covered by~$x$\;
  Let ${S(x)\subseteq \{1,\dots,n\}}$  denote
  the vertices that are incident on the edges in~$U(x)$\;
  Choose $b \in \{0, 1\}$ uniform at random\;
  \If{$b=0$ \Or $S(x)=\emptyset$} 
  {flip each bit of $x$ independently with 
    probability~$1/n$}
  \Else{    
    flip each bit of $S(x)$ independently with probability~$1/2$\;\label{li:hi-mut}
    flip each bit of $\{1,\ldots,n\} \setminus S(x)$ independently with probability~$1/n$
  }
\end{algorithm2e}

The results presented rely on an alternative mutation operator (see Algorithm~\ref{alg:paramcomplex:ch-cover-altmutation}) that has the ability to perform bit flips with a high probability if the corresponding node is adjacent to at least one uncovered edge (line \ref{li:hi-mut} of Algorithm~\ref{alg:paramcomplex:ch-cover-altmutation}). This allows the algorithm to perform random sampling on the subgraph consisting of the uncovered edges. 
If this subgraph constitutes a kernel of the problem, the random sampling process is similar to a brute-force search on the kernel. We will summarize those results in the following. 

We outline the results for the algorithms introduced in this section, but should also mention that the vertex cover problem has been subject to further parameterized analyses in the context of randomized search heuristics.
For example, the investigations of the vertex cover problem that we present in this section have been extended to the weighted vertex cover problem~\cite{PPSN2016WeightedVCP}. 
Gao et al.~\cite{DBLP:conf/ppsn/GaoFN16} have studied random initialization heuristics as well as local search algorithms in terms of parameterized complexity and approximation.  Furthermore, the vertex cover problem has been analyzed in dynamic settings where edges can be removed from or added to the graph~\cite{UsDVCGecco2015}. 

\subsection{Parameterized Analysis}

The first parameterized result in the context of optimal vertex covers considers \gsemoalt together with the objective function $f_1$, which uses the number of uncovered edges as the second objective. The population size of the algorithm is upper bounded by $n+1$, as the main objective (number of chosen nodes) can only take on that many different values. The same upper bound on the population size is applied when using $f_2$.

The first analysis relies on the following basic insight. Let $\OPT$ be the value of an optimal solution; then an optimal solution has to include all nodes of degree at least $\OPT +1$. This is based on the simple observation that if a node $v$ of degree $\OPT+1$ is not selected, all neighbors of $v$ have to be selected, resulting in a nonoptimal solution.

\begin{theorem}
\label{thm:paramcomplex:cover-gsemoalt}
The expected optimization time of \gsemoalt for  
the minimum vertex cover problem using the fitness function $f_1$ is upper bounded by $O(\OPT\cdot n^4+ n\cdot 2^{\OPT+\OPT^2})$.
\end{theorem}

The proof of the theorem proceeds in several different phases. First, the expected time until the search point $0^n$ is included in the population is analyzed. The proof for this part focuses on selecting the individual with the smallest number of $1$-bits, which happens with probability at least $1/(n+1)$, as the number of different values for $|x|_1$ is at most $n+1$. Producing a solution with a smaller number of $1$-bits is always accepted, and the problem can be seen as maximizing the number of $0$-bits, slowed down by a population of size at most $n+1$. Hence, after an expected number of $O(n^2 \log n)$ steps of \gsemo or \gsemoalt using $f_1$ or $f_2$, the search point $0^n$ is included in the population. 

We now consider $f_1$ and assume that the search point $0^n$ is already included in the population. 
Subsequently, the expected number of steps where the population does not contain a solution $x$ for $f_1$ that is a kernel for the problem is upper bounded by $O(\OPT \cdot n^4)$. For $f_1$, $x$ is a kernel iff the vertices chosen by $x$ constitute a subset of an optimal solution and the maximum degree of $G(x)$ is at most $\OPT$. In order to upper bound the number of steps where the population does not contain a solution $x$ that is a kernel, a potential function with $O(n^2 \OPT)$ different values is taken into account that measures the population with respect to the number of uncovered edges that its individuals have. It can be shown that the potential can always be improved with probability at least $\Omega(1/n^2)$ if no kernel is contained in the population. As the potential cannot increase, the expected number of steps where the population does not contain a kernel is $O(n^4 \cdot \OPT)$

Denoting by $\hat{x}$ the resulting vertex cover, the kernel instance $G(\hat{x})$ has at most $\OPT^2 + \OPT$ nonisolated nodes. In this case, the alternative mutation operator is able to produce the optimal solution from $\hat{x}$ in expected time $O(n\cdot 2^{\OPT+\OPT^2})$. In this upper bound, the factor $n$ accounts for selecting the individual $\hat{x}$ with probability at least $1/(n+1)$ and the term $O(2^{\OPT+\OPT^2})$ accounts for mutating this individual into an optimal solution. The exponential component of the runtime arises from the waiting time to make a lucky random jump, but this jump is now required only on a reasonably small kernel instance.

The runtime bound can be improved if the value of an optimal linear program $LP(x)$ for the graph $G(x)$ consisting only of the uncovered edges is used as the second criterion, leading to the fitness function $f_2$. The goal is to minimize the penalty $LP(x)$, and we have $LP(x)=0$ iff $x$ is a vertex cover.

The analysis is based on the following result of
Nemhauser and Trotter~\cite{Nemhauser1975}, who proved a very strong relation between optimal fractional vertex covers and minimum vertex covers.

\begin{theorem}\label{thm:paramcomplex:NT}
Let~$x^*$ be an optimal fractional vertex cover and let~$P_0,P_1\subseteq V$ be the vertices whose corresponding components of~$x^*$ are~$0$ or~$1$, respectively. Then there exists a minimum vertex cover that contains~$P_1$ and no vertex of~$P_0$.
\end{theorem}

Theorem~\ref{thm:paramcomplex:NT} implies that one can take all vertices set to $1$ in an optimal fractional vertex cover and reduce the size of the problem in this way. Furthermore, it is well known that every basic feasible solution $x$ of the vertex cover LP relaxation is half-integral, i.e., we have $x \in \{0,1/2,1\}^n$~\cite{Balinski1970}.  Using these properties, the following result has been shown.

\begin{theorem}\label{thm:paramcomplex:lptimeopt}
The expected optimization time of \gsemoalt for  
the minimum vertex cover problem using the fitness function $f_2$ is upper bounded by~$O(n^2\cdot\log n+\OPT\cdot n^2+n\cdot 4^{\OPT})$.
\end{theorem}

We now explain the key ideas of the proof.
We already know that the population contains the search point $0^n$ after an expected number of $O(n^2 \log n)$ steps.
After $0^n$ has been included in the population, the number of steps where the population does not contain a kernel is investigated. For $f_2$, a solution $x$ is a kernel iff $LP(x) = LP(0^n)-|x|_1$ and each optimal fractional vertex cover assigns $1/2$ to each nonisolated vertex of $G(x)$. The number of steps where $P$ does not contain such a kernel $x$ after $0^n$ has been included in the population can be bounded by $O(\OPT \cdot n^2)$ using the following arguments. Solutions with objective value $(r, LP(0^n)-r)$ are Pareto optimal. The proof proceeds by considering the solution $x$ with objective vector $(r, LP(0^n)-r)$ and the largest value of $r$ in the population. If $x$ is not a kernel, that $x$ can be chosen for mutation with a probability of at least $1/(n+1)$ and one specific bit can be flipped with a probability of at least $1/(en)$ to produce a Pareto-optimal offspring $x'$ with objective vector $(r+1, LP(0^n)-r-1)$. As the value of the LP is upper bounded by $\OPT$, at most $\OPT$ of such steps can happen. This upper bounds the number of additional steps (after $0^n$ has been included in the population) by $O(n^2 \cdot \OPT)$.

Let $\hat{x}$ be the kernel with objective vector $(r, LP(0^n)-r)$, where $r$ is the maximum such that all nonisolated vertices of $G(x)$ obtain a value of $1/2$ in $LP(\hat{x})$. $G(\hat{x})$ has at most $2(\OPT-|\hat{x}|_1) \leq 2 \cdot \OPT$ nonisolated vertices, as the vertices that are chosen belong to an optimal solution and every nonisolated vertex contributes $1/2$ to the LP value. The expected time to produce an optimal solution after a kernel $\hat{x}$ has been included in the population is $O(n \cdot 2^{2\cdot \OPT}) = O(n \cdot 4^{\OPT})$, as the optimal solution can be obtained by choosing $\hat{x}$ for mutation and flipping exactly the bits corresponding to the nonisolated nodes of an optimal solution while not flipping the remaining bits.

Kratsch and Neumann have also given the following trade-off results with respect to runtime and approximation. These results show the previous FPT time bound ($\epsilon=0$), as well as that \gsemoalt achieves a $2$-approximation ($\epsilon=1$) in expected polynomial time.

\begin{theorem}\label{thm:paramcomplex:lptimeepsilonapprox}
Using the fitness function $f_2$, the expected number of iterations of \gsemoalt  until it has generated a~$(1+\epsilon)$-approximate vertex cover, i.e., a solution of fitness~$(r,0)$ with~$r\leq(1+\epsilon)\cdot \OPT$, is~$O(n^2\cdot\log n+\OPT\cdot n^2+n\cdot 4^{(1-\epsilon)\cdot \OPT})$.
\end{theorem}

The proof of Theorem~\ref{thm:paramcomplex:lptimeepsilonapprox} uses the same kernelization arguments as the proof of Theorem~\ref{thm:paramcomplex:lptimeopt}.
Once a solution $\hat{x}$ that is a kernel of the problem has been produced, it is shown that if $\hat{x}$ is selected for mutation then it will mutate  with probability $\Omega((1/4)^{(1-\epsilon)\cdot \OPT'})$ into a solution $x'$ for which

$$|x'|_1+2\cdot LP(x') \leq (1+\epsilon)\cdot \OPT$$
holds. Such a solution $x'$ can be turned into a vertex cover by single mutation steps that reduce $LP(x)$ by at least $1/2$ while increasing the size of the vertex cover by one, leading to a vertex cover of size at most  $(1+\epsilon)\cdot \OPT$.

\section{Submodular Functions with Constraints}
\label{sec:paramcomplex:sub}

Submodular functions constitute a broad class of interesting problems. 
A function $f \colon 2^X \rightarrow \R$ is submodular iff
$f(A \cup B) + f(A \cap B) \leq f(A) + f(B)$ for all $A, B \subseteq X$. In the context of optimizing a submodular function $f$, we will often consider
the incremental value of adding a single element, leading to an equivalent definition.
We denote by $F_i(A)= f(A \cup \{i\}) - f(A)$
the marginal value of $i$ with respect to $A$.
A function $f$ is submodular
iff $F_i(A) \geq F_i(B)$ for all $A\subseteq B \subseteq X$
and $i\in X \setminus B$.

We consider the problem of maximizing a given submodular function $f$. The problem is \textsf{NP}-hard, as it generalizes
many \textsf{NP}-hard combinatorial optimization problems, such as
maximum cut~\cite{GoemansW95,FeigeG95} and several others~\cite{AgeevS99,Cornuejols1977,Hastad01,FeigeG95},
The class of submodular functions also includes the class of linear functions that have been well studied in the area of theory of evolutionary computation. Friedrich and Neumann~\cite{DBLP:journals/ec/FriedrichN15} have analyzed the maximization of submodular functions with different constraints and carried out runtime analyses depending on the parameters of the given constraint. We will summarize the results in this section.

Friedrich and Neumann considered the maximization of a given submodular function $f$ under a given set of matroid constraints.
A matroid is a pair $(X,\cI)$ composed of a ground set $X$ and a nonempty collection $\cI$ of subsets of $X$
satisfying
(1) if $A\in\cI$ and $B \subseteq A$ then $B \in \cI$ and,
(2) if $A, B \in \cI$ and $|A| > |B|$ then $B + x \in \cI$ for some $x \in A \setminus B$.
The sets in $\cI$ are called \emph{independent},
and the \emph{rank} of a matroid is the size of any maximal independent set. We will consider several different classes of submodular functions together with different types of matroid constraints.

Friedrich and Neumann analyzed the \oea and \gsemo as baseline algorithms. For the \oea,  the fitness function
$h(x)= (v(x), f(x))$ was considered. Here, $v(x)$ measures the constraint violation of $x$.
Generalizing the fitness function used by Reichel and Skutella~\cite{ReichelSkutella10}
for the intersection of two matroids, they
considered problems with $k$ matroid constraints $M_1, \ldots, M_k$,
$$v(x) = k \cdot |x|_1 - \sum_{j=1}^k r_j(x),$$
where $r_j(x)$ denotes the rank of $x$ in matroid $M_j$, i.e., 
$$r_j(X) = \max \{ |Y| \colon Y \subseteq X, Y \in I_j\}$$ for the set $X$ given by $x$.

We have $v(x)=0$ iff $x$ is a feasible solution and $v(x)>0$ otherwise.  
The function $h(x)$ is optimized in lexicographic order, i.e.,
$$h(y) \geq h(x) \text{ holds iff } (v(y) < v(x)) \vee (v(y)=v(x) \wedge f(y) \geq f(x)).$$
We denote by $F$ the set of feasible solutions. For \gsemo, Friedrich and Neumann set  $z(x) = f(x)$ iff $x \in F$ and $z(x) =-1$ iff $x \not \in F$ and considered the multiobjective problem
$g(x) := (z(x), |x|_0),$
where $|x|_0= \sum_{i=1}^n (1-x_i)$ denotes the number of $0$-bits in the given bit string $x$. Adding the number of $0$-bits as the second objective to be maximized forces the empty set to be Pareto optimal, and allows the algorithm to construct solutions greedily.

\subsection{Monotone Functions with Uniform Constraints}
We now summarize the results for the special class of monotone submodular functions under one uniform matroid constraint.
 A function $f$
is monotone iff
$f(A) \leq f(B)$ for all $A\subseteq B$. A uniform matroid constraint of size $r$ means that a set is feasible iff it consists of at most $r$ elements, i.e., $\cI = \{A \subseteq X \colon |A| \leq r\}$.

A key property of \gsemo that is often employed in theoretical analysis is that it constructs solutions in a manner similar to a greedy algorithm. 
Furthermore, the population size can be bounded by $n+1$, as the number of different objective values for the second objective is $n+1$. This implies that one particular individual that is needed for the analysis is selected with probability $\Omega(1/n)$.
The algorithm removes elements in order to maximize the number of zeros. Using the number of zeros as the second objective implies that the algorithm maintains a population where the solution with the smallest number of elements is never removed. Furthermore, each solution that has a smaller number of selected elements than the solutions previously found is included in the population. Eventually, this leads to a population which includes the solution consisting of the empty set.
 In terms of the first objective (the overall goal function), the algorithm tries to maximize its objective value in a greedy manner. It does so by adding elements that provide the largest benefit to a current solution.  Putting these arguments together, the following approximation result can be obtained for \gsemo and the maximization of monotone submodular functions with a uniform constraint.

\begin{theorem}
\label{thm:paramcomplex:uniform}
The expected time until \gsemo has obtained a $(1- 1/e)$-approximation for a monotone submodular function $f$ under a uniform constraint of size $r$ is $O(n^2 \,(\log n+r))$.
\end{theorem}

The proof of the theorem uses the fact that the population size is always bounded by $n+1$ and therefore one particular individual is selected with probability at least $1/(n+1)$ in each step. The first phase of the proof shows that the empty set, represented by the bit string $0^n$, is included in the population in expected time $O(n^2 \log n)$. Similarly to the analysis for vertex cover in the previous section, this bound is obtained by considering the factor $O(n)$ for the population size and bounds on a coupon collector process for maximizing the number of $0$-bits. The $O(n^2r)$ term accounts for the greedy process where the correct individual in the population is selected with probability $\Omega(1/n)$ and the appropriate greedy step is applied to this individual with probability $\Omega(1/n)$. Finally, there are at most $r$ of these steps, as no more than $r$ elements can be inserted owing to the given constraint. The approximation ratio follows from the greedy process.

\subsection{Monotone Submodular Functions under Matroid Constraints}

Now we take a look at more complex problems. Again we consider monotone submodular functions but with $k$ matroid constraints.
The algorithm that we consider is the \oea. The number of these matroid constraints is the important parameter that we consider and it determines the approximation ratio that is achieved, as well as the exponent of the runtime. Furthermore, there is a parameter $p\geq 1$ that allows for a fixed value of $k$ to trade off the approximation quality and runtime of the algorithm.

\begin{theorem}
\label{thm:paramcomplex:oneonepex}
For any integers $k\geq 2$, $p\geq 1$ and a real value $\epsilon>0$, the expected time until the \oea has obtained a $(1/(k+ 1/p + \epsilon))$-approximation for any monotone submodular function $f$ under $k$ matroid constraints is $O{\left(\frac{1}{\epsilon} \cdot n^{2p(k+1)+1} \cdot k \cdot \log n\right)}$.
\end{theorem}

We summarize the main ideas of the proof here. The first part of the proof consists of showing that the algorithm reaches a feasible solution $x$ with $f(x) \geq \OPT/n$. The expected time until the \oea has obtained such a solution can be upper bounded by $O(n^{k+1})$. To attain this bound, the proof first argues that the \oea obtains a feasible solution in expected time $O(kn \,( \log k + \log n))$ by using the fitness level method applied to the value of the penalty $v(x)$. Afterwards, it is shown that, from any feasible solution $x$, a feasible solution $y$ with $f(x) \geq \OPT/n$ can be obtained by flipping $k+1$ specific bits. The expected waiting time for this event is $O(n^{k+1})$.

 A $p$-exchange operation applied to the current solution $x$ introduces at most $2p$ new elements and deletes at most $2kp$ elements of $x$. 
A solution $y$ that can be obtained from $x$ by a $p$-exchange operation is called a $p$-exchange neighbor of $x$.
According to \cite{LeeSV10}, every solution $x$ for which there exists no $p$-exchange neighbor $y$ with $f(y) \geq (1+ \frac{\epsilon}{n(k+1)})\cdot f(x)$ is a $(1/(k+1/p +\epsilon))$-approximation for any monotone submodular function. So, the proof works by analyzing the time until a feasible solution has been obtained. Afterwards, it uses the fact that there is still a $p$-exchange neighbor unless the desired approximation quality has already been obtained.

\subsection{Symmetric Submodular Functions under Matroid Constraints}

We now summarize the main result for \gsemo for the optimization of symmetric submodular functions under $k$ matroid constraints. The following theorem makes use of the greedy and local search ability that the algorithm \gsemo has.

\begin{theorem}
\label{thm:paramcomplex:semolee}
The expected number of iterations until \gsemo attains a $\left(\frac{1}{(k+2) (1+\epsilon)} \right)$-approximation
for any symmetric submodular function under $k$ matroid constraints is $O{\left(\frac{1}{\epsilon} n^{k+6} \log n\right)}$, for any constant $\epsilon>0$.
\end{theorem}

The analysis makes use of the following result in~\cite{Lee2009}, which shows that there are always locally improving steps as long as the desired approximation quality has not been obtained.
\begin{lemma}
\label{lem:paramcomplex:approx}
Let $x$ be a solution such that no solution with fitness at least $\left(1+ \frac{\epsilon}{n^4}\right) \cdot f(x)$ can be achieved by deleting one element or by inserting one element and deleting at most $k$ elements. Then $x$ is a $\left(\frac{1}{(k+2) (1+ \epsilon)} \right)$-approximation.
\end{lemma}

The proof of Theorem~\ref{thm:paramcomplex:semolee} uses this lemma together with the fact that \gsemo introduces the search point $0^n$ into the population after an expected number of $O(n^2 \log n)$ steps. As the search point $0^n$ is Pareto optimal, it stays in the population once it has been introduced. Selecting $0^n$ for mutation and inserting the element that leads to the largest increase in the $f$-value produces a solution $y$ with $f(y) \geq \OPT/n$. The reason for this is that the number of elements is limited by $n$ and that $f$ is submodular. \gsemo will also always have a solution with the largest $f$-value obtained so far in the population. Selecting this solution $x$ for mutation and flipping at most $k+1$ specific bits according to Lemma~\ref{lem:paramcomplex:approx} produces a solution $y$ with
$f(y) \geq  \left(1+ \frac{\epsilon}{n^4}\right) \cdot f(x)$
as long as $x$ does not yet have the desired approximation quality.
The expected waiting time for this event is $O(n^{k+2})$, as at most $k+1$ specific bits of $x$ have to be flipped and the population size is at most $n+1$.

The number of steps that improve the solution with the largest $f$-value needed in order to achieve the desired $\left(\frac{1}{(k+2) (1+ \epsilon)} \right)$-approximation  is upper bounded by 
$$\log_{\left(1+ \frac{\epsilon}{n^4}\right)} \frac{\OPT}{\OPT/n} = O\bigg(\frac{1}{\epsilon}\, n^4 \log n\bigg)$$
which implies that the expected time to achieve a $\left(\frac{1}{(k+2) (1+ \epsilon)} \right)$-approximation is $O{\left(\frac{1}{\epsilon} n^{k+6} \log n\right)}$.

\section{Euclidean TSP}

\label{sec:paramcomplex:tsp}

Given a set of $n$ points $V = \{v_1, v_2, \ldots, v_n\}$ in the plane, the objective of the Euclidean TSP is to find a permutation $\pi\colon V \to V$ that minimizes the cost function
\begin{equation}
c(\pi) = \sum_{i=1}^n d(v_{\pi(i)},v_{\pi(i+1)}),
\label{eq:paramcomplex:tsp-cost}
\end{equation}
where $d(v_i, v_j)$ denotes the Euclidean distance separating the points $v_i$ and $v_j$ and arithmetic is taken to be modulo $n$. The Euclidean TSP is \textsf{NP}-hard, but can be approximated to within a factor $(1+\epsilon)$ for every fixed $\epsilon$ in polynomial time~\cite{DBLP:journals/jacm/Arora98}.

It is convenient to consider the complete undirected graph $G = (V,E)$ and define the Hamiltonian cycle $C(\pi) \subseteq E$ induced by the edges followed by a given permutation $\pi$:
\[
C(\pi) = \{ \{v_{\pi(1)},v_{\pi(2)}\}, \{v_{\pi(2)}, v_{\pi(3)}\}, \ldots, 
\{v_{\pi(n-1)},v_{\pi(n)}\}, \{v_{\pi(n)},v_{\pi(1)}\} \}.
\]
We will refer to the cycle $C(\pi)$ as a \emph{tour}.

Iterative improvement methods rely on the iterated exchange of a small number of edges and are powerful approaches for solving large-scale TSP instances in practice.  These heuristics move through the space of candidate solutions by repeatedly applying move or mutation operators to pivot between tours. For the TSP, this is typically some variant of the powerful $k$-opt operation. The $k$-opt move considers some candidate tour $C(\pi)$, and deletes $k$ mutually disjoint edges and reassembles the remaining fragments into a new valid tour $C(\pi')$. The operation induces a neighborhood structure on the search space of tours, and thus serves as a strong and easy-to-implement local search operator. However, instances exist where this approach is provably inefficient. 
For example, local search algorithms employing a $k$-opt neighborhood operator can take exponential time even to find a locally optimal solution~\cite{Chandra1999new}. This even holds for the Euclidean case~\cite{Englert2007worst}.

The convex hull of $V$ is the smallest convex set containing $V$. A point $v \in V$ is called an \emph{inner point} if $v$ lies in the interior of the convex hull of $V$. We denote by $\Inn{V} \subset V$ the set of inner points of $V$, and define $\Out{V} := V \setminus \Inn{V}$. The TSP parameterized by $k = \Inn{V}$ is in $\textsf{FPT}$. Specifically, De{\u{\i}}neko et al.~\cite{Deineko2006inner} showed that if a Euclidean TSP instance with $n$ vertices has $k$ vertices interior to the convex hull, there is a dynamic programming FPT algorithm. Other parameterizations are not as propitious; for example, finding a local optimum in the $k$-opt neighborhood for the metric TSP is hard for $\textsf{W}[1]$~\cite{Marx2008tsp}. $\textsf{FPT} \subseteq \textsf{W}[1]$, but the containment is conjectured to be proper~\cite{Flum2006parameterized}, in which case no such FPT algorithm can exist. 

Parameterized results for evolutionary algorithms for the Euclidean TSP have been developed in a series of papers~\cite{Sutton2012tsp,CEC2013,DBLP:conf/cec/NallaperumaSN13a,DBLP:journals/ec/SuttonNN14} in the context of the inner-point parameterization of De{\u{\i}}neko et al.~\cite{Deineko2006inner}. We also would like to mention that the generalized traveling salesperson problem has been investigated in the context of parameterized complexity. In this problem, the cities belong to different clusters and the goal is to compute a shortest tour that visits each cluster exactly once. We refer the interested reader for details of the generalized TSP to Corus et al.~\cite{DBLP:journals/ec/CorusLNP16}.

 The remainder of this section sketches these results, starting with the setting in which the algorithm is oblivious to problem-specific information (other than the cost of a tour) and ending with algorithms that exploit problem-specific structure. 

\subsection{Black-Box Algorithms} 
\label{sec:paramcomplex:tsp:black-box}
In the black-box setting, heuristics are not allowed any access to domain-specific knowledge about the instance other than the cost of a tour. For Euclidean TSP instances with $k = \Inn{V}$ inner points, it is possible to show that the \mplea generates an optimal solution in slicewise polynomial time (that is, in time $n^{g(k)}$, where $g$ depends only on $k$). Later, in Section~\ref{sec:paramcomplex:tsp:fpt}, we will discuss how it is possible to improve this to FPT time when domain knowledge is incorporated into the design of the algorithm.

\begin{figure} 

  \centering

  \begin{tikzpicture}[ 
    scale=0.3,
    ActionEdge/.style={very thick},
    Edge/.style={},
    PathSegment/.style={
      very thick, 
      dashed, 
      decoration={snake,segment length=1cm, amplitude=0.2cm},
      decorate
    },
    Vertex/.style={fill=white}
    ]
          
    %% Set up coordinates for vertices
    \foreach \i in {0,...,7}
    \coordinate (c\i) at ({5*cos(deg(2*\i*pi/8))},{5*sin(deg(2*\i*pi/8))});

    %% Label the coordinates and address them
    \def\names{x1,xim1,xj,xjm1,xip1,xi,xjp1,xn};
    \foreach \name [count=\i] in \names
    {      
      \pgfmathparse{Mod(4-\i,8)}
      \coordinate (\name) at (c\pgfmathresult);
    }
    
    % Draw edges
    \draw[ActionEdge] (xim1) edge (xi);
    \draw[Edge] (xi) edge (xip1);
    \draw[Edge] (xjm1) edge (xj);
    \draw[ActionEdge] (xj) edge (xjp1);

    \draw[Edge] (xn) edge (x1);

    %% Draw path segments 
    \draw[PathSegment] (x1) -- (xim1); 
    \draw[PathSegment] (xip1) -- (xjm1); 
    \draw[PathSegment] (xjp1) -- (xn); 

    % Draw vertices
    \foreach \i in {0,...,7}
    \draw[Vertex] (c\i) circle (0.5cm);

    % Label vertices
    \node[draw=none,above=0.2cm,anchor=south] at (xim1) {$x(i-1)$};
    \node[draw=none,above right=0.2cm,anchor=west] at (xj) {$x(j)$};
    \node[draw=none,right=0.2cm,anchor=west] at (xjm1) {$x(j-1)$};
        
    \node[draw=none,below left=0.2cm,anchor=east] at (xjp1) {$x(j+1)$};
    \node[draw=none,below=0.2cm,anchor=north] at (xi) {$x(i)$};
    \node[draw=none,below right=0.2cm,anchor=west] at (xip1) {$x(i+1)$};
        
    \node[draw=none,above left=0.2cm,anchor=east] at (x1) {$x(1)$};
    \node[draw=none,left=0.2cm,anchor=east] at (xn) {$x(n)$};               
  \end{tikzpicture}
  \begin{tikzpicture}[ 
    scale=0.3,
    ActionEdge/.style={very thick},
    Edge/.style={},
    PathSegment/.style={
      very thick, 
      dashed, 
      decoration={snake,segment length=1cm, amplitude=0.2cm},
      decorate
    },
    Vertex/.style={fill=white}
    ]
          
    %% Set up coordinates for vertices
    \foreach \i in {0,...,7}
    \coordinate (c\i) at ({5*cos(deg(2*\i*pi/8))},{5*sin(deg(2*\i*pi/8))});

    %% Label the coordinates and address them
    \def\names{x1,xim1,xj,xjm1,xip1,xi,xjp1,xn};
    \foreach \name [count=\i] in \names
    {      
      \pgfmathparse{Mod(4-\i,8)}
      \coordinate (\name) at (c\pgfmathresult);
    }
    
    % Draw edges
    \draw[ActionEdge] (xim1) edge (xj);
    \draw[ActionEdge] (xi) edge (xjp1);
    \draw[Edge] (xjm1) edge (xj);
    \draw[Edge] (xi) edge (xip1);

    \draw[Edge] (xn) edge (x1);

    %% Draw path segments 
    \draw[PathSegment] (x1) -- (xim1); 
    \draw[PathSegment] (xip1) -- (xjm1); 
    \draw[PathSegment] (xjp1) -- (xn); 

    % Draw vertices
    \foreach \i in {0,...,7}
    \draw[Vertex] (c\i) circle (0.5cm);

    % Label vertices
    \node[draw=none,above=0.2cm,anchor=south] at (xim1) {$x(i-1)$};
    \node[draw=none,above right=0.2cm,anchor=west] at (xj) {$x(j)$};
    \node[draw=none,right=0.2cm,anchor=west] at (xjm1) {$x(j-1)$};
        
    \node[draw=none,below left=0.2cm,anchor=east] at (xjp1) {$x(j+1)$};
    \node[draw=none,below=0.2cm,anchor=north] at (xi) {$x(i)$};
    \node[draw=none,below right=0.2cm,anchor=west] at (xip1) {$x(i+1)$};
        
    \node[draw=none,above left=0.2cm,anchor=east] at (x1) {$x(1)$};
    \node[draw=none,left=0.2cm,anchor=east] at (xn) {$x(n)$};               
  \end{tikzpicture}

  \caption{\label{fig:paramcomplex:inversion} The effect of the inversion operation
    $\inv{i}{j}$ on a tour. Inverting a subsequence in the
    permutation representation corresponds to a 2-opt move in which a
    pair of edges in the current tour is replaced by a
    pair of edges not in the tour. }
\end{figure}
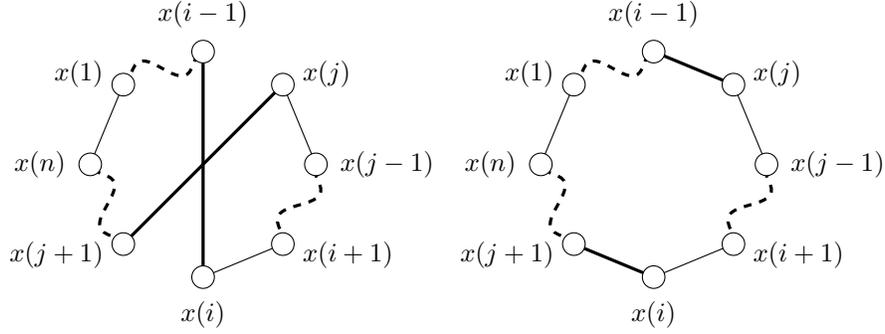

The 2-opt operator mentioned above corresponds to segment reversal in the linear form of the corresponding tour permutation. We refer to the 2-opt operation as the \emph{inversion} operation and illustrate it in Fig.~\ref{fig:paramcomplex:inversion}. We consider random local search (RLS), defined in Algorithm~\ref{alg:paramcomplex:rls}, and the \mplea, defined in Algorithm~\ref{alg:paramcomplex:ea}. Note that RLS maintains a population of size one, and performs exactly one inversion operation in each iteration. On the other hand, the \mplea maintains a population of $\mu$ permutations and produces $\lambda$ offspring in each generation by applying Poisson mutation (see Function~\FuncSty{\ref{fun:paramcomplex:mutate}}).

\begin{definition}
  \label{def:paramcomplex:inversion}
  The \emph{inversion} operation $\inv{i}{j}$ transforms permutations
  into one another by segment reversal in their linear forms.

  A permutation $x$ is transformed into a permutation $\inv{i}{j}[x]$
  by inverting the subsequence of the linear form of $x$ from position $i$ to position $j$, where $1 \leq i < j \leq n$:
  \begin{align*}  
    x &= (x(1), \ldots, x(i-1), x(i), x(i+1),\ldots,x(j-1),x(j),x(j+1), \ldots, x(n)),\\
    \inv{i}{j}[x] &= (x(1), \ldots, x(i-1), x(j),
    x(j-1),\ldots,x(i+1),x(i),x(j+1), \ldots, x(n)).
  \end{align*}
\end{definition}

We also consider the permutation \emph{jump} operator studied by Scharnow, Tinnefeld, and
Wegener~\cite{STWsorting} in the context of sorting problems.

\begin{definition}
  \label{def:paramcomplex:jump}
  The \emph{jump} operation $\jmp{i}{j}$ transforms permutations into
  one another by position shifts in their linear form. A permutation $x$ is transformed
  into a permutation $\jmp{i}{j}[x]$ by moving the element in position
  $i$ in the linear form of $x$ into position $j$ in the linear form of $\jmp{i}{j}[x]$ while the other elements between position $i$
  and position $j$ are shifted in the appropriate direction.  Without
  loss of generality, suppose $i < j$. Then,
  \begin{align*}  
    x &= (x(1), \ldots, x(i-1), x(i), x(i+1),\ldots,x(j-1),x(j),x(j+1), \ldots, x(n)),\\
    \jmp{i}{j}[x] &= 
    (x(1), \ldots, x(i-1), x(i+1),\ldots,x(j-1), x(j), x(i),x(j+1), \ldots, x(n)).
  \end{align*}
\end{definition}

\begin{algorithm2e}
  \caption{\label{alg:paramcomplex:rls}Randomized local search (RLS)}  
  \SetKwFor{For}{repeat}{}{}
  Choose a random permutation $x$ on $V$\;
  \For{forever}{%
    choose a random distinct pair of elements $(i,j)$ from $[n]$\;
    $y \gets \inv{i}{j}[x]$\;
    \lIf{$f(y) \leq f(x)$}{$x \gets y$}
  }
\end{algorithm2e}

\begin{function}
\SetProcNameSty{texttt}
  \caption{mutate($x$)} \label{fun:paramcomplex:mutate}
$y \gets x$\;
draw $s$ from a Poisson distribution with unit expectation\;
perform $s+1$ random inversion operations on $y$\;
\Return $y$\;
\end{function}

\begin{algorithm2e}
  \caption{\label{alg:paramcomplex:ea}The \mplea}  
  \SetKwFunction{mutate}{mutate}
  \SetKwFunction{select}{select}
  \SetKwFor{For}{repeat}{}{}
  Choose a multiset $P$ of $\mu$ random permutations on $V$\;
  \For{forever}{%
    $P' \gets \{\}$\;
    \For{$\lambda$ times}{%
      choose $x$ uniformly at random from $P$\;\label{li:select-rep}
      $y \gets \mutate(x)$\;
      $P' \gets P' \uplus \{y\}$\;
    }
    $P \gets \select(P \uplus P')$ \;
  }
\end{algorithm2e}

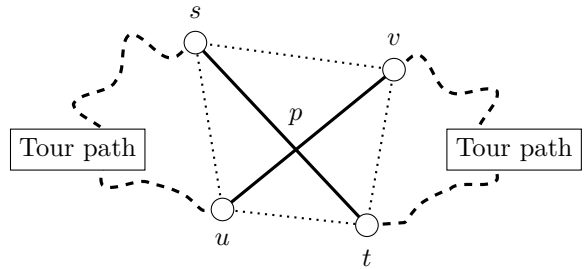
\begin{figure}
  \centering  
  \begin{tikzpicture}[ 
    scale=0.3,
    ActionEdge/.style={very thick},
    Edge/.style={},
    PathSegment/.style={
      very thick, 
      dashed, 
      decoration={snake,segment length=1cm, amplitude=0.2cm},
      decorate
    },
    Vertex/.style={fill=white}
    ]
          
  \coordinate (u) at (0.2,0.8);
  \coordinate (s) at (-1,8.2);
  \coordinate (t) at (6.6,0.1);
  \coordinate (v) at (7.8,7);  

  \draw[ActionEdge,name path=Edge1] (s) -- (t);
  \draw[ActionEdge,name path=Edge2] (u) -- (v);
  \path [name intersections={of=Edge1 and Edge2,by=p}];

  \draw[PathSegment] (s) to[out=180,in=180,looseness=2] (u);
  \draw[PathSegment] (v) to[out=0,in=0,looseness=2] (t);

  \draw[dotted,thick] (s) -- (v) -- (t) -- (u) -- cycle;

  \draw[Vertex] (u) circle (0.5cm);
  \node[draw=none,below=0.2cm,anchor=north] at (u) {$u$};

  \draw[Vertex] (s) circle (0.5cm);
  \node[draw=none,above=0.2cm,anchor=south] at (s) {$s$};

  \draw[Vertex] (t) circle (0.5cm);
  \node[draw=none,below=0.2cm,anchor=north] at (t) {$t$};

  \draw[Vertex] (v) circle (0.5cm);
  \node[draw=none,above=0.2cm,anchor=south] at (v) {$v$};

  \node[above=0.2cm,anchor=south] at (p) {$p$};  

  \node[left=2cm of p,draw,fill=white] {Tour path};
  \node[right=2cm of p,draw,fill=white] {Tour path};

  \end{tikzpicture}
  \caption{Removing the intersecting edges $(s,t)$ and $(u,v)$ and reconnecting the two disconnected tour path segments with edges $(s,v)$ and $(u,t)$ results in a strictly shorter tour.}
  \label{fig:paramcomplex:intersect}
\end{figure}

Every tour $C(\pi)$, for all permutations $\pi$ on $V$, corresponds to a set of edges that describe a closed polygon in the plane. If $V$ is noncollinear (no three points are collinear), the vertices on the boundary of the convex hull of $V$ appear in their cyclic order in a minimum-cost tour, and no edge is intersecting~\cite{Quintas1965properties}. 
When a tour contains a pair of edges that intersect at a point $p$, those edges form the diagonals of a convex quadrilateral. The interior edges of this figure describe nondegenerate triangles in the Euclidean plane. Thus, as long as no three points are collinear, removing these edges and replacing them with the corresponding nonintersecting edges results in a strictly shorter tour. This is illustrated in Fig.~\ref{fig:paramcomplex:intersect}.

\subsubsection{Avoiding Arbitrarily Small Improvements}
Worst-case proofs for 2-opt on the TSP exploit the fact that when points are allowed in arbitrary positions, the smallest change in fitness between neighboring solutions can be made arbitrarily small~\cite{Englert2007worst}. This allows the possibility of exponential-length paths between a candidate solution and a reachable local optimum. Sutton and Neumann~\cite{Sutton2012tsp} circumvented this is by imposing bounds on the angles between points. 
A set of points $V$ is \emph{angle-bounded by $\epsilon$} for some $0 < \epsilon < \pi/2$ if, for any three
points $u,v,w \in V$, $0 < \epsilon < \theta < \pi-\epsilon$, where
$\theta$ denotes the angle formed by the line from $u$ to $v$ and the
line from $v$ to $w$. Under this condition, the runtime bound depends on the angle bound $\epsilon$, and so we may consider it as an additional parameterization of the instance. This is also applicable to the class of TSP instances whose points are embedded in an $m \times m$ grid (with the further restriction that no three points are collinear). This kind of quantization can result when the coordinates of each point are rounded to the nearest value in a set of $m$ equidistant values. In these cases, the changes in cost between neighboring solutions can be bounded from below, avoiding exponentially long improvement chains to reach a local optimum.

\begin{definition}\label{def:paramcomplex:fn}
  Let $V$ be a set of points angle-bounded by $\epsilon$. We define
  \[  
  \fn(\epsilon) = \left(\frac{d_{\rm max}}{d_{\rm min}} -
    1\right)\left(\frac{\cos(\epsilon)}{1-\cos(\epsilon)}\right)
  \]
  where $d_{\rm max}$ and $d_{\rm min}$ denote the maximum and
  minimum Euclidean distances, respectively, between points in $V$.
\end{definition}

Quantized instances yield a more meaningful interpretation of $\fn(\epsilon)$, as is captured by the following proposition.
\begin{proposition}
  \label{prp:quantized}
  Let $V$ be a set of points embedded in an $m \times m$ grid with no three points collinear. Then $V$ is angle-bounded by $\epsilon$ such that
  \[
  A(\epsilon) = m^5.
  \]
\end{proposition}

Proposition~\ref{prp:quantized} follows from Definition~\ref{def:paramcomplex:fn} and the fact that $V$ is angle-bounded by $\arctan
\left(1/(2(m-2)^2)\right)$ and $d_{\rm max} = O(m)$.

\subsubsection{Instances in Convex Position}
A set of points $V$ are in convex position when $\Inn{V} = \emptyset$. In this case, we must wait only for the process to remove all intersecting edges. Upper bounds on the time until RLS and the \mplea have removed all such edges (and thus produced an optimal tour) can be expressed as a function of the angle-bounding function $\fn$. More conveniently, when an instance is embedded in an $m \times m$ grid, both processes can solve the instance in time polynomial in both $n$ and $m$.

\begin{theorem}
  \label{thm:paramcomplex:convex-position-rls}
  Let $V$ be a set of planar points in convex position angle-bounded
  by $\epsilon$. The expected time for RLS to solve the TSP on $V$ is
  $O(n^3 \fn(\epsilon))$, where $\fn$ is as defined in
  Definition~\ref{def:paramcomplex:fn}.
 \end{theorem}

The proof of Theorem~\ref{thm:paramcomplex:convex-position-rls} relies on the fact that any 2-opt move that replaces a pair of intersecting edges with a pair of nonintersecting edges in an angle-bounded instance results in an improvement of the tour by at least 
\begin{equation}
2 d_{\rm min} \left(1-\cos(\epsilon)\right)/\left(\cos(\epsilon)\right).\label{eq:paramcomplex:rls-c}    
\end{equation}
Any pair of intersecting edges can be removed with a particular 2-opt operation (each of which occurs with probability $\Omega(n^{-2})$), and thus we can derive a straightforward bound on the waiting time until all such intersections have been removed.

\begin{theorem}
  \label{thm:paramcomplex:convex-position-mulambda}
  Let $V$ be a set of planar points in convex position angle-bounded
  by $\epsilon$. The expected number of fitness evaluations needed by the \mplea using 2-opt mutation to solve the TSP on $V$ is bounded from above by
  $O{\left(n\cdot\fn(\epsilon)\cdot\max\left\{\mu n^2,\lambda\right\}\right)}$, where $\fn$ is as defined in
  Definition~\ref{def:paramcomplex:fn}.
\end{theorem}

The proof of Theorem~\ref{thm:paramcomplex:convex-position-mulambda} is similar to the proof of Theorem~\ref{thm:paramcomplex:convex-position-rls}, except that we must account for any slowdown incurred by selecting from a population. Specifically, the probability that at least one of the $\lambda$ offspring improves on the current best-so-far point is at least $1 - \left(1 - \frac{1}{\mu e n(n-1)/2}\right)^\lambda$. When $\lambda \geq \mu n(n-1)/2$, an intersection is removed with constant probability in each generation and we must wait only $O(n\fn(\epsilon))$ generations to find an intersection-free tour (owing to the improvement guarantee from \eqref{eq:paramcomplex:rls-c}). On the other hand, when $\lambda < \mu e n (n-1)/2$, the improvement probability can be as low as $\lambda/(\mu e n^2)$. The runtime bound follows by accounting for this and the extra $\mu+\lambda$ fitness evaluations that need to occur in each generation. 

\subsubsection{Bounded Number of Inner Points}
The polynomial-time results on angle-bounded instances in convex position raise the question of what kind of influence the number of inner points can have on the running time of the above-mentioned algorithms. In this section, we discuss how the Euclidean TSP parameterized by the number of inner points can be solved in \emph{slicewise polynomial time} in the black-box setting.

\begin{theorem}
  \label{thm:paramcomplex:ea-2opt}
  Let $V$ be a set of points angle-bounded by $\epsilon$ such that
  $|\Inn{V}| = k$.  The expected number of fitness evaluations needed for the
  \mplea using 2-opt mutation to solve the TSP on $V$ is
  bounded from above by 
  \[
  O{\left(n\cdot\fn(\epsilon)\cdot\max\left\{\mu n^2,\lambda\right\} +
      \mu n^{4k}(2k-1)!\right)},
  \] 
  and the expected optimization time for the \oea is
  \[
  O{\left(n^3\cdot\fn(\epsilon) + n^{4k}(2k-1)!\right)}.
  \]
 \end{theorem}

Theorem~\ref{thm:paramcomplex:ea-2opt} can be proved by partitioning the amount of time the \mplea spends on tours that contain intersections and tours that do not contain intersections. In particular, let $x^{(t)}$ be the best-so-far tour found by generation $t$ of the \mplea. If $C(x^{(t)})$ contains a pair of intersecting edges, the probability of the EA creating a strictly improving tour via a 2-opt mutation on $x^{(t)}$ is bounded from below. Moreover, the angle-boundedness of the instance guarantees an additional lower bound on the amount of actual fitness improvement when such a mutation occurs. Hence, the total expected time that the process spends on tours with intersecting edges is bounded as in Theorem~\ref{thm:paramcomplex:convex-position-mulambda}.

In the case where $x^{(t)}$ contains no intersecting edges, the vertices on the boundary of the convex hull must appear in $x^{(t)}$ in their correct cyclic order for a minimum-cost tour~\cite{Quintas1965properties}. An optimal tour can then be produced from $x^{(t)}$ by rearranging the points in $\Inn{V}$ to the correct positions. Poisson mutation (see Function~\FuncSty{\ref{fun:paramcomplex:mutate}}) is capable of performing this rearrangement by selecting at most $2|\Inn{V}| = 2k$ specific inversion operations. This occurs with probability at least
\[
\frac{1}{e n^{4k}(2k-1)!},
\]
which yields a simple upper bound on the waiting time to jump from an intersection-free tour to an optimal solution. The claim then follows by carefully accounting for the correct parent selection probabilities and summing the bounds on the expected time spent on tours with intersections and nonoptimal intersection-free tours.

\subsubsection{Mixed-Mutation Strategies}
The proofs of the theorems in the preceding sections rely on the inversion operator to construct an intersection-free tour, but then rely on the inversion operator to simulate a jump operation in order to transform the intersection-free tour into an optimal solution. The analysis can be improved by relying on a mixed-mutation strategy (see Function~\FuncSty{\ref{fun:paramcomplex:mixed-mutation}}) that performs a mixture of both inversion and jump operations, each with constant probability. This improves the upper bound on the running time by a factor of $\Omega{\left(n^{2k}(2k-1)!/(k-1)!\right)}$.

\begin{function}
\SetProcNameSty{texttt}
  \caption{mixed-mutation($x$)} \label{fun:paramcomplex:mixed-mutation}
$y \gets x$\;
draw $r$ from a uniform distribution on the interval $[0,1]$\; 
draw $s$ from a Poisson distribution with unit expectation\;
\lIf{$r < 1/2$}{perform $s+1$ random inversion operations on $y$}
\lElse{perform $s+1$ random jump operations on $y$}
\Return $y$\;
\end{function}

\begin{theorem}
  \label{thm:paramcomplex:ea-mixed1}
  Let $V$ be a set of points angle-bounded by $\epsilon$ such that $|\Inn{V}| = k$.  The expected number of fitness evaluations needed for the
  \mplea using mixed mutation to solve the TSP on $V$ is
  bounded from above by 
  \[
  O{\left(n\cdot\fn(\epsilon)\cdot\max\left\{\mu n^2,\lambda\right\} +
      \mu n^{2k}(k-1)!\right)},
  \]
 and the expected optimization time for the \oea is bounded from above by 
 \[
 O{\left(n^3\cdot\fn(\epsilon) + n^{2k}(k-1)!\right)}.
 \]
\end{theorem}

The proof is similar to the proof of Theorem~\ref{thm:paramcomplex:ea-2opt}. With mixed mutation, a 2-opt operation still occurs with constant probability, so the likelihood of a sufficient improvement is asymptotically equivalent to the case of Theorem~\ref{thm:paramcomplex:ea-2opt}. A jump operation occurs also with constant probability, but the probability that such an operation jumps to an optimal solution (by correctly rearranging the positions of the points in $\Inn{V}$) is bounded from below by
\[
\Omega{\left(\frac{1}{n^{2k}(k-1)!}\right)}.
\]

\subsection{FPT Evolutionary Algorithms}
\label{sec:paramcomplex:tsp:fpt}
In the case where search heuristics have access to problem-specific information, FPT results are also available. Specifically, we consider heuristics that have access to both fitness values and the cyclic ordering of the points on the convex hull. This ordering can be precomputed in polynomial time~\cite{DBLP:journals/ipl/Graham72} and stored so that it is available to the heuristic at any time.

\subsubsection{A Population-Based Approach}
\label{sec:paramcomplex:tsp:fpt:pop}
Building on a previous study of Theile~\cite{Theile2009exact}, Sutton et al.~\cite{DBLP:journals/ec/SuttonNN14} constructed a population-based evolutionary algorithm that efficiently solves the Euclidean TSP when the number of inner points is not too large. They showed that a small modification to Theile's \mpoea that carefully maintains the invariant that the points in $\Out{V}$ remain in correct convex-hull order for each individual results in an FPT evolutionary algorithm for the inner-point parameterization of the Euclidean TSP.

The EA maintains a large population of permutations on subtours in the graph $G = (V,E)$ (a \emph{subtour} is a Hamiltonian cycle on a subset of $V$). In each generation, a new offspring is created via a specialized mutation operator that extends the subtour by incorporating an additional randomly chosen vertex, and a modified truncation selection is applied that chooses the best individual for a subtour. The EA can be seen as an evolutionary approach to dynamic programming, the framework for which was presented in~\cite{DynEA}.

For a set of $n$ points $V$ in the plane with $|\Inn{V}| = k$, we denote by $\gamma := (p_1, p_2, \ldots,p_{n-k})$ a linear order on the points of $\Out{V}$ such that for all $i \in \{1,\ldots, n-k\}$, $p_i$ and $p_{i+1}$ are adjacent on the boundary of the convex hull of $V$. For any subset $U \subseteq V$, a permutation on $U$ is a bijection $x\colon U \to U$. We say that a permutation $x$ on $U \subseteq V$ is $\gamma$-respecting if and only if, for all $p_i, p_j \in U$, $x^{-1}(p_i) < x^{-1}(p_j) \implies i < j$. We call $U$ the \emph{ground set} of the permutation $x$ on $U$. We refer to the first element $x(1)$ in the linear order of such a permutation as the \emph{head vertex} and the last element $x(|U|)$ as the \emph{tail vertex}. 

The \mplea maintains a population $P$ of $\gamma$-respecting permutations on subsets of $V$. For each subset $S \subseteq \Inn{V}$ and each $i \in [n-k]$, the population $P$ contains permutations on the ground set $S \cup \{p_1, p_2, \ldots, p_i\}$. There are $(|S|+i)!$ possible permutation on this ground set. If we were to allow all of them in the population, $|P|$ would be exponential in $n$. Hence, the key to the FPT running time of the EA is the realization that in an optimal solution, the points in $\Out{V}$ must always appear in their order around the hull. Therefore it is wasteful to consider permutations that are not $\gamma$-respecting. 

\begin{figure}
  \centering
  \begin{tikzpicture}[
    x=2cm,y=2cm,
    decoration={markings,
      mark=at position 0.42 with {\coordinate (anchor);}}
    ]
    \tikzstyle{vertex}=[draw,circle,minimum size=6mm,fill=white]
    \tikzstyle{edge}=[very thick]
            
    \coordinate (p1) at (1.1,-1);
    \coordinate (p2) at (0.3,0);
    \coordinate (p3) at (0.8,1);
    \coordinate (p4) at (2.3,1.1);
    \coordinate (p5) at (3,0.2);
    \coordinate (p6) at (2.6,-0.9);
    \coordinate (s1) at (1.2,0.5);
    \coordinate (s2) at (0.9,-0.2);
    \coordinate (s3) at (2,-0.5);
    \coordinate (s4) at (2.1,0.3);
    \coordinate (s5) at (1.5,0);

    \draw[dashed,fill=gray!30,postaction=decorate] plot[smooth cycle] coordinates {(p1) (p2) (p3) (p4) (p5) (p6) (p1)};

    \node[above right=5mm of anchor,inner sep=1pt,outer sep=0mm] (label) {Convex hull};
    \draw[-latex] (label.south west) -- (anchor);

    \node[vertex] (np1) at (p1) {$p_1$}; 
    \node[vertex] (np2) at (p2) {$p_2$}; 
    \node[vertex] (np3) at (p3) {$p_3$};
    \node[vertex] (np4) at (p4) {$p_4$}; 
    \node[vertex] (np5) at (p5) {$p_5$};
    \node[vertex] (np6) at (p6) {$p_6$};
    \node[vertex] (ns1) at (s1) {$u$}; 
    \node[vertex] (ns2) at (s2) {$v$}; 
    \node[vertex] (ns3) at (s3) {};
    \node[vertex] (ns4) at (s4) {$r$};
    \node[vertex] (ns5) at (s5) {};
    
    \draw[edge] (np1) -- (ns1);
    \draw[edge] (ns1) -- (np2);
    \draw[edge] (np2) -- (ns2);
    \draw[edge] (ns2) -- (np3);
    \draw[edge] (np3) -- (np4);
    \draw[edge] (np4) -- (ns4);
    \draw[edge] (ns4) -- (np1);
    
  \end{tikzpicture}
  \caption{The subtour defined by the permutation $x_{(i,S,r)} = (p_1,u,p_2,v,p_3,p_4,r)$ where $S = \{u,v,r\}$ and $i=4$. The positions of the points $p_i \in \Out{V}$ in the linear order of the permutation respect their cyclic order around the convex hull.}
  \label{fig:paramcomplex:subtour}
\end{figure}
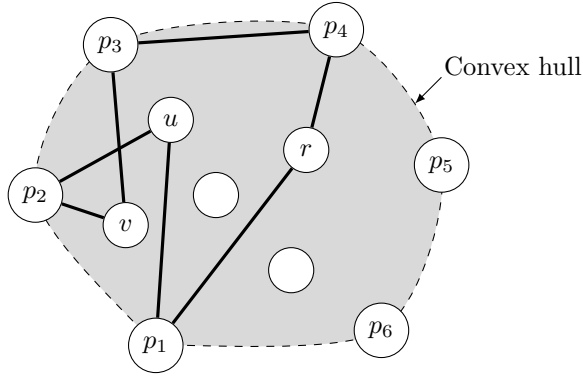

To exploit this, for each possible ground set $S \cup \{p_1, p_2, \ldots, p_i\}$, the population contains exactly $|S|+1$ $\gamma$-respecting permutations on that ground set, one for each possible unique tail vertex from the ground set. Specifically, for every $S \subseteq \Inn{V}$ and every $i \in [n-k]$ there is a permutation $x$ for every $r \in S \cup \{p_i\}$ such that
\begin{enumerate}
  \item the head vertex of $x$ is $x(1) = p_1$,
  \item the tail vertex of $x$ is $x(|S| + i) = r$, and
  \item $x$ is $\gamma$-respecting.
\end{enumerate}
We denote a permutation over the ground set $S \cup \{p_1, p_2, \ldots, p_i\}$ with tail vertex $r$ by $x_{(i,S,r)}$. The corresponding subtour of a $x_{(i,S,r)}$ is a cycle $(x(1) = p_1, v_{x(2)}, \ldots, v_{x(|S|+i-1)}, r, p_1)$ that starts at $p_1$ and runs through each point of the ground set $U$ exactly once (the $i$ points of $\Out{V}$ are visited in the order in which they appear in $\gamma$). Finally, the cycle visits $r$ before returning to $p_1$. An illustration of a subtour for an example permutation $x_{(i,S,r)}$ on a small ground set is depicted in Fig.~\ref{fig:paramcomplex:subtour}. The fitness function utilized by the \mplea is simply the cost of the subtour of an individual:
\begin{equation}
  \label{eq:paramcomplex:fitness2}
  f(x_{(i,S,r)}) = \sum_{j=1}^{|S| + i} d(v_{x(j)},v_{x(j+1)}),
\end{equation}
where the summation indices are taken to be modulo $|S| + i$.

For any given $S \subseteq \Inn{V}$, there are $n-k$ ways to construct a ground set (by choosing $i$) and $|S|+1$ ways to choose the tail vertex from $S \cup \{p_i\}$. The total number of individuals in the population is thus
\[
\mu = |P| = (n-k) \sum_{s=0}^k \binom{k}{s}(s+1) = O(2^kkn).
\]

The specially designed mutation operator extends a permutation $x = x_{(i,S,r)}$ by adding exactly one new point to its ground set, preserving the validity constraints. In particular, a vertex $v$ is chosen uniformly at random from the remaining vertices in $(\Inn{V} \setminus S) \cup \{p_{i+1}\}$.\footnote{We have abused notation slightly by taking $\{p_{|\Out{V}|+1}\}$ to mean $\emptyset$.} 
A new permutation $x'$ is constructed from $x$ by concatenating $v$ with the linear order described by $x$; that is, for $j \in \{1, \ldots, |S| + i+1\}$,
\[
x'(j) = \begin{cases}
  v & \text{~if $j = |S|+i+1$,}\\
  x(j) & \text{~otherwise.}\\
\end{cases}
\]
Thus $x'$ is a permutation over the ground set $S \cup r$ and uses $v$ as the new tail vertex:
\[
x' = 
\begin{cases}
  x'_{(i,S \cup \{v\},v)} & \text{~if $v \in \Inn{V}$,}\\
  x'_{(i+1, S,v)} & \text{~if $v = p_{i+1}$.}\\
\end{cases}
\]
When $i = n-k$ and $S = \Inn{V}$, the mutation operator has no effect, since the ground set cannot be extended for such an individual.

In each generation of the \mplea, $\lambda$ individuals are selected uniformly at random from $P$. For each selected individual $x$, an offspring is generated by composing the mutation operator described above $s+1$ times, where $s$ is drawn from a Poisson distribution with unit expectation. Survival selection proceeds by ensuring that each mutated offspring may replace only the individual in the parent population with the same ground set and tail vertex, and this replacement occurs only when the fitness of the offspring is at least as good as the fitness of the corresponding parent. In this way, the surviving population maintains the invariant that each valid combination of ground set and tail vertex is represented exactly once. 

\begin{algorithm2e}[t]
  \caption{\label{alg:paramcomplex:mu+lambda} \mplea}
  \SetKwFor{For}{repeat}{}{}
  $P \gets \emptyset$\; \label{li:initstart}
  \ForEach{$ i \in \{1, \ldots, n-k\}$}{
    \ForEach{$S \subseteq \Inn{V}$}{
      \ForEach{$r \in S \cup p_i$}{
        $x \gets$ a permutation on the ground set $S \cup \{p_1, p_2, \ldots, p_i\}$ such that $x(|S|+i) = r$ and $x$ respects $\gamma$\;
        $P \gets P \cup x$\;
      }
    }
  }\label{li:initend}
  
   \For{forever}{%
    $P' \gets \{\}$\;
  	\For{$\lambda$   times}
  	{
    Select an individual $z \gets x_{(i,S,r)} \in P$ uniformly at random\;

    Draw $s$ from a Poisson distribution with unit expectation\;\label{li:pois}
    Generate $z' \gets x_{(i',S',r')}$ by applying the mutation operator $s+1$ times\;
    Let $f(z')$ be the cost of TSP tour generated by $z'$\; 
    $P' \gets P' \cup z'$\;
    }
    \tcc{truncation selection based on the same ground set}
    \ForEach{offspring $z'$ in $P'$}{
    Let $z'' \gets x_{(i',S'r')} \in P$ be an individual defined on the same ground set as $z'$ having the same end vertex if such an individual exists in the population\;
    \lIf { $f(z') \leq f(z'')$ } {$P \gets P \cup z' \setminus z''$}
    }
  }
\end{algorithm2e}

\begin{theorem}
  \label{thm:paramcomplex:dp-fpt}
  Let $V$ be a set of $n$ points in the Euclidean plane with $|\Inn{V}| = k$. After $O(\max\{2^k k^2n^{2} \lambda^{-1},n\})$ generations, the \mplea solves the TSP on $V$ to optimality in expectation and with probability $1 - e^{-\Omega(n)}$.
\end{theorem}

Note that this bound translates to $O(\max\{2^k k^2n^2, \lambda n\})$ fitness evaluations in expectation, by taking the random numbers counting fitness evaluations and generations to be $T_f$ and $T_g$, respectively, and noting that for Algorithm~\ref{alg:paramcomplex:mu+lambda}, $E[T_f] = \mu + \lambda E[T_g]$. The proof of Theorem~\ref{thm:paramcomplex:dp-fpt} proceeds by bounding the time it takes to increase the set of optimal subtours in the population. In particular, we say that a population is \emph{solved to order $m$} when it contains an individual permutation on a ground set of size $m$ that corresponds to an optimal subtour on that ground set. Obviously, such subtours are never lost (since they cannot be replaced by a suboptimal subtour), and the initial population is solved to order $1$ since it contains the individual $x_{(p_1,\emptyset,p_1)}$. The claim follows by bounding the probability of a transformation from a population solved to order $m$ to one solved to order $m+1$, and subsequently taking the waiting time to get a population solved to order $n$.

\subsubsection{Inner-Point Permutations}
As we saw in Section~\ref{sec:paramcomplex:tsp:fpt:pop}, incorporating domain knowledge into the design of an EA can allow us to create a randomized FPT algorithm for a particular parameterization of the Euclidean TSP. Algorithm~\ref{alg:paramcomplex:mu+lambda}, however, potentially needs a large population, specifically $\mu = O(2^k kn)$.
Another approach is to keep a small population and use an EA to search for the optimal ordering on the inner points. Specifically, we let $\gamma = (p_1, p_2, \ldots, p_{n-k})$ be the fixed order of points in $\Out{V}$ as they appear on the convex hull. For any permutation $x\colon \Inn{V} \to \Inn{V}$, it is straightforward to compute the value of the optimal tour through $\Inn{V}$ and $\Out{V}$ respecting the order of both $\gamma$ and $x$. The naive approach is to try all $O(n^k)$ possible ways of merging the linear orders of the permutations $\gamma$ and $x$. This would violate our FPT requirement, since the parameter appears in the power of the polynomial. Instead, to preserve our FPT conditions, we can directly use a dynamic programming approach to compute the fitness of the permutation $x$ on $\Inn{V}$. 

We define two $(n-k) \times (k+1)$ matrices $F^{\rm Out}$ and $F^{\rm Inn}$, where $F^{\rm Out}[i,j]$ (or $F^{\rm Inn}[i,j]$) stores the value of the minimum-weight subtour of all tours through points $p_1, p_2, \ldots, p_i$ and $x(1), x(2), \ldots, x(j)$ such that they respect the orders of both $\gamma$ and $x$, and they end on an outer point (or inner point, respectively). Then the optimal tour given the permutations $\gamma$ and $x$ is
\begin{eqnarray*}
  Dyn(x)  =  \min  \{ F^{\rm Out}[n-k,k] + d(p_{n-k}, p_1),
  F^{\rm Inn}[n-k,k]+ d(x(k),p_1) \}.
\end{eqnarray*}
Taking the boundary case as $F^{\rm Out}[1,0]=0$ (the subtour consisting only of $p_1$), we can compute
\begin{eqnarray*}
  F^{\rm Inn}[i,j]  =  \min  \{ F^{\rm Out}[i,j-1] + d(p_{i}, x(j)), F^{\rm Inn}[i,j-1]+ d(x(j-1),x(j)) \}
\end{eqnarray*}
for $i \in \{1,2,\ldots,n-k\}$ and $j \in \{1,\ldots,k\}$, and
\begin{eqnarray*}
  F^{\rm Out}[i,j]  =  \min  \{ F^{\rm Out}[i-1,j] + d(p_{i-1}, p_i),
  F^{\rm Inn}[i-1,j]+ d(x(j),p_i) \}
\end{eqnarray*}
for $i \in \{2,3,\ldots,n-k\}$ and $j \in \{0,\ldots,k\}$. Entries that do not correspond to valid subtours, namely $F^{\rm Out}[1,j]$ for $j \geq 1$ (since the tour cannot end on $p_1$ and then return to $p_1$) and $F^{\rm Inn}[i,0]$ for $i \geq 1$ (since a subtour cannot end on an inner point when the inner-point set is empty), are set to $\infty$.

The two $F$ matrices can be computed in $O(nk)$ time using dynamic programming. Thus, the time complexity of the fitness evaluation of $Dyn(x)$ is $O(nk)$. 

\begin{algorithm2e}
  \caption{\label{alg:paramcomplex:eak}\mpleak}
   \SetKwFunction{select}{select}
  \SetKwFor{For}{repeat}{}{}
  Choose a multiset $P$ of $\mu$ random permutations on $V$\;
  \For{forever}{%
    $P' \gets \{\}$\;
    \For{$\lambda$ times}{%
      Choose $x$ uniformly at random from $P$\;
      Draw $s$ from a Poisson distribution with unit expectation\;
      Construct $x'$ from $x$ by applying $s+1$ random basic operations\;
    Let $f(x')$ be $ Dyn(x')$\;
    $P' \gets P' \cup x'$\;
  }
   $P \gets \select(P \uplus P')$ \; 
  }
\end{algorithm2e}

\begin{theorem}
  \label{thm:paramcomplex:eak-fpt}
Let $V$ be a set of $n$ points in the Euclidean plane with $|\Inn{V}| = k$. Assuming $\lambda = O(\mu)$, the \mpleak solves the TSP on $V$ using at most $O(\mu + (k-1)!k^{2k})$ fitness evaluations with the jump operation as the basic mutation operation. This bound can be improved to $O(\mu + (k-2)!k^{2k-2})$ by using 2-opt mutation. Moreover, each fitness evaluation has time complexity $O(nk)$.
\end{theorem}

Note that we state the theorem slightly differently than in~\cite{DBLP:journals/ec/SuttonNN14}, in which the expected number of \emph{generations} was proved to be $O(\max\{(k-1)!k^{2k}\lambda^{-1},1\})$ for jumps and $O(\max\{(k-2)!k^{2k-2}\lambda^{-1},1\})$ for 2-opt mutation. The bounds stated in Theorem~\ref{thm:paramcomplex:eak-fpt} follow by noting that the number of fitness evaluations in $T_g$ generations of Algorithm~\ref{alg:paramcomplex:eak} is $\mu + \lambda T_g$, and the added assumption about $\lambda$. The proof of Theorem~\ref{thm:paramcomplex:eak-fpt} relies again on the probability that a given mutation correctly arranges the inner points. Since the mutation operation performs $s+1$ random basic operations, where $s$ is Poisson distributed, the probability that it performs $\ell$ basic operations is $e^{-1}/(\ell - 1)!$. On a permutation of length $k$, a distinct jump (or 2-opt) move is chosen uniformly at random with probability at least $k^{-2}$, so the probability that a \emph{specific sequence} of $\ell$ basic operations occurs is at least
\[
p(k,\ell) = \frac{1}{e(\ell-1)! k^{2\ell}}.
\]
Therefore, the waiting time to create a globally optimal offspring is bounded by the diameter of the search space induced by the mutation operator. For 2-opt, this bound is at most $k-1$~\cite{Bafna1996genome}, and for the jump operation, the bound is $k$. In the case of jump, the probability that at least one of the $\lambda$ offspring created in any generation is optimal is at least $1 - (1- p(k,k))^\lambda \geq \min\{\lambda p(k,k), 1 - e^{-1}\}$. The claim follows from a standard waiting-time argument. We improve the bound for 2-opt by substituting $p(k,k-1)$ in the above transformation probability.

\section{Conclusion}
\label{sec:paramcomplex:conclusion}
In this chapter, we have presented an outline of recent results on the parameterized complexity analysis of randomized search heuristics. This approach of incorporating additional salient parameters into running-time analysis allows a finer-grained understanding of the influence of problem structure on the behavior of these general-purpose optimization techniques.

We have seen that a parameterized analysis can illuminate the inherent efficiency of particular search operators, as well as reveal the difficult components that might arise in the search space of a problem instance. This is the case for the maximum-leaf spanning tree problem. On graphs where $k$ is the maximum number of leaves in a spanning tree, a tree-preserving mutation operator guarantees that the \oea can find such a tree in 
fixed-parameter tractable time $O(2^{15k^2 \log k})$. This is in contrast to standard mutation, for which there exist graphs with $m$ edges requiring $(m/c)^{\Omega(k)}$ steps.

We have also observed that the concept of kernelization from the theory of parameterized complexity can be useful. Multiobjective algorithms using a specialized mutation operator can focus the search on a problem kernel of the vertex cover problem, leading to an FPT running time. We have explored how parameterized analysis can help to strengthen an understanding of the components of very general problem classes on simple evolutionary algorithms. This is the case, for example, with the maximization of submodular functions under different constraints.

For the Euclidean TSP, the inner-point parameterization of De{\u{\i}}neko et al.~\cite{Deineko2006inner} illuminates the difficulty for RSH techniques arising from the number of points that lie inside the convex hull of the instance. This informs the design of FPT problem-specific evolutionary algorithms, but so far the best known black-box analysis for this parameterization remains in XP time. An open problem is therefore either to prove that this is a lower bound for the parameterization, or to improve the upper bound to FPT time. 

Traditional running-time analyses of randomized search heuristics on some artificial benchmark functions have already implicitly used a parameterized perspective. One clear example is for the \textsc{Jump} function, the running time analysis of which is typically parameterized by the jump-gap size ($k$) and the string length ($n$). Indeed, the running-time dichotomy between mutation-only evolutionary algorithms ($\Omega(n^k)$~\cite{Jansen2002Crossover}) and recombinant evolutionary algorithms ($O(4^k \poly(n))$~\cite{Jansen2002Crossover,KotzingST:c:11:crossover}) already exhibits an ``FPT-like'' flavor. The application of parameterized analysis to running-time analysis of randomized search heuristics on combinatorial optimization problems with well-established parameterizations from the classical community is therefore a very natural research direction.

Perhaps the most significant research requirement is the need for good problem parameterizations. This requires theoreticians to work closely with practitioners in order to understand what problem components are the most meaningful and relevant in the real world, i.e., what features are most likely to be manifested (or be restricted) in practice, and what problem characteristics might be exploitable by different techniques. This emphasizes the importance of a strong and vibrant relationship between theory and practice.

\end{document}